\documentclass{article}

\PassOptionsToPackage{numbers, compress}{natbib}
\usepackage{arxiv}

\usepackage[utf8]{inputenc} 
\usepackage[T1]{fontenc}    
\usepackage{hyperref}       
\usepackage{url}            
\usepackage{booktabs}       
\usepackage{amsfonts}       
\usepackage{nicefrac}       
\usepackage{microtype}      
\usepackage{xcolor}         
\usepackage{graphicx}
\usepackage{multirow}
\usepackage{float}
\usepackage{subfig}
\usepackage{amsmath}
\usepackage{paralist}
\usepackage{todonotes}
\makeatletter
\newcommand*\iftodonotes{\if@todonotes@disabled\expandafter\@secondoftwo\else\expandafter\@firstoftwo\fi}  
\makeatother

\usepackage{wrapfig}
\usepackage[capitalize]{cleveref}

\title{Pathologies in Priors and Inference\\for Bayesian Transformers}

\date{}

\author{%
    Tristan Cinquin \\
    ETH Zürich\\
    Zürich, Switzerland \\
    \texttt{tcinquin@student.ethz.ch} \\
    \And
    Alexander Immer \\
    ETH Zürich \\
    Zürich, Switzerland \\
    \texttt{alexander.immer@inf.ethz.ch} \\
    \AND
    Max Horn \\
    ETH Zürich \\
    Basel, Switzerland \\
    \texttt{max.horn@bsse.ethz.ch} \\
    \And
    Vincent Fortuin \\
    ETH Zürich \\
    Zürich, Switzerland \\
    \texttt{fortuin@inf.ethz.ch} \\
}

\begin{document}

\maketitle

\begin{abstract}

In recent years, the transformer has established itself as a workhorse in many applications ranging from natural language processing to reinforcement learning.
Similarly, Bayesian deep learning has become the gold-standard for uncertainty estimation in safety-critical applications, where robustness and calibration are crucial.
Surprisingly, no successful attempts to improve transformer models in terms of predictive uncertainty using Bayesian inference exist.
In this work, we study this curiously underpopulated area of Bayesian transformers.
We find that weight-space inference in transformers does not work well, regardless of the approximate posterior.
We also find that the prior is at least partially at fault, but that it is very hard to find well-specified weight priors for these models.
We hypothesize that these problems stem from the complexity of obtaining a meaningful mapping from weight-space to function-space distributions in the transformer.
Therefore, moving closer to function-space, we propose a novel method based on the implicit reparameterization of the Dirichlet distribution to apply variational inference directly to the attention weights.
We find that this proposed method performs competitively with our baselines.

\end{abstract}

\section{Introduction}

The transformer \citep{vaswani2017attention} is a deep learning architecture commonly used to process sequences of data, such as text. Thanks to multi-head self-attention, the transformer builds contextual embeddings by capturing the relationships between the sequence elements. While being most famous for their state-of-the-art performance in natural language processing \citep{brown2020gpt3, devlin2019bert}, transformers are also used in computer vision \citep{chen2020imagegpt, dosovitskiy2021vit, jiang2021transgan, strudel2021segmenter, wu2020visualtransformers}, reinforcement learning \citep{chen2021decisiontransformer, kumar2020adaptive}, as well as audio \citep{gong2021ast, huang2018musictransformer, payne2019musenet} and video \citep{yan2021videogpt} processing, yielding impressive results.
    
The Bayesian learning paradigm provides a theoretical framework to obtain predictive uncertainty, select the optimal model, and improve its calibration. 
Furthermore, by designing an informative prior for the parameters, Bayesian models offer a principled way to incorporate assumptions about the inferred distribution, thus providing regularization. Finally, recent work \citep{kristiadi2020bayesian, mitros2019validity} has shown that Bayesian neural networks (BNN) are often better calibrated than standard neural networks.

If transformers and Bayesian deep learning are both so popular, why have we not seen any successful Bayesian transformer models? By attempting to implement such models, we make the following contributions: (i) We find that weight space inference in transformers does not provide any improvements over a model trained by maximum likelihood. (ii) We show that the prior is at least partially at fault for this. (iii) We propose to perform inference on the attention weights rather than on the parameters, and present a novel variational method for this using the Dirichlet distribution.  

\section{Background}
\vspace{-0.75em}

\subsection{Bayesian deep learning}

Bayesian inference computes the posterior distribution as
\begin{equation}
    \mathit{P}(\theta \mid y_{1:N}, x_{1:N}) = \mathit{P}(y_{1:N} \mid \theta, x_{1:N}) \mathit{P}(\theta) / \mathit{P}(y_{1:N} \mid x_{1:N})
    \label{eq:posterior}
\end{equation}
with neural network parameters $\theta$, training data $\{(x_i, y_i)\}_{i=1}^N$, likelihood function $\mathit{P}(y_{1:N} \mid \theta, x_{1:N})$, prior $\mathit{P}(\theta)$, and evidence $\mathit{P}(y_{1:N} \mid x_{1:N})$.
The predictive distribution of a new target $y^*$ given $x^*$ is then obtained by
\begin{equation}
    P(y^* \mid x^*, y_{1:N}, x_{1:N}) = \mathrm{E}_{
        \theta \sim \mathit{P}(\theta \mid y_{1:N}, x_{1:N})} 
            [\mathit{P}(y^* \mid \theta, x^*)]
    \label{eq:predicitve}
\end{equation}
Applied to neural networks, both \cref{eq:posterior} and \cref{eq:predicitve} are intractable and need to be estimated using approximate inference methods, such as variational inference or Monte Carlo sampling.

\subsection{Bayesian neural network weight space inference}

\begin{wrapfigure}{r}{0.5\linewidth}
\vspace{-10pt}
\includegraphics[width=\linewidth]{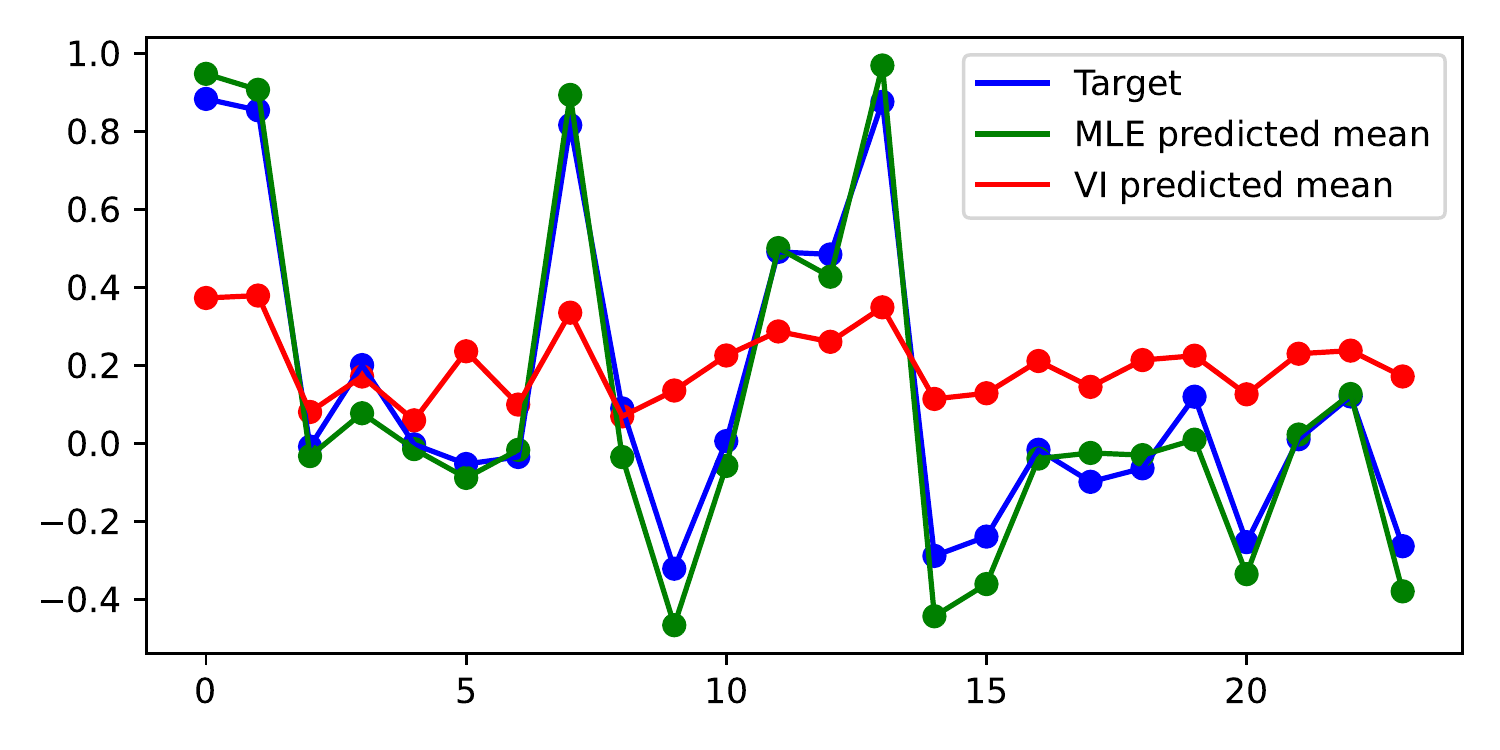}
\caption{Plot of MLE and VI transformer. MLE captures the mean of the generative process well (MSE: 0.66), but VI does not (MSE: 9.58).}
\label{fig:m3_vi_plots}
\vspace{-10pt}
\end{wrapfigure}

The most commonly used weight space inference methods in BNNs are variational inference (VI) \citep{blundell2015weight, dusenberry2020efficient, gal2017concretedropout, louizos2016maxtrixgaussianposteriors, louizos2017multiplicativenfvi, mishkin2019slang}, the Laplace method \citep{daxberger2021bayesian, immer2021glm, kristiadi2020bayesian}, and Markov Chain Monte Carlo (MCMC) \citep{chen2014stochastichamiltonianmcmc, neal2011hamiltonianmcmc, welling2011sgld}.
While MCMC methods directly sample from the (unnormalized) posterior, VI and Laplace approximate the posterior by another distribution. As MCMC methods are expensive computationally and memory-wise, we restrict our focus to VI and Laplace. 
When applying these methods to transformers, we find that weight space inference fails to improve data fit, calibration and predictive uncertainty compared to a model trained by likelihood maximization (see~\cref{fig:m3_vi_plots}).
 
\subsection{Empirical weight study}

To understand why weight space inference fails, we study the empirical weight distribution of transformers trained with stochastic gradient descent (SGD), hoping to obtain better priors. We follow the framework proposed by \citet{fortuin2021bayesian}.
We first examine the marginal weight distribution where we especially study the tailedness and modality. We also identify the best-fitting distribution and its parameters within the Gaussian, Student, Logistic, Cauchy, and Laplace families.
Furthermore, we investigate the correlation among layer weights by comparing the empirical covariance matrix and the distribution of off-diagonal covariance elements against samples from an isotropic Gaussian.

\section{Methods}
\vspace{-0.75em}

\subsection{Variational attention}

As as alternative to weight-space inference in transformers, we propose to treat self-attention weights as random variables and approximate their posterior distribution using VI. Previous attention weight inference methods focus on sampling \citep{an2020repulsive}, while others explicitly parameterize the attention weights with a particular distribution \citep{bahuleyan2018variationalattention, deng2018latentalignment, fan2020bam}. 
Parameters of explicitly reparameterizable distributions such as the Gaussian \cite{bahuleyan2018variationalattention}, Weibull, and Lognormal distributions \citep{fan2020bam} are learned via VI, while others such as the Dirichlet \citep{deng2018latentalignment} require using REINFORCE gradient estimators \citep{sutton2000reinforce}.


We implement two baselines for our comparison:
\begin{inparaenum}
\item \textit{Gaussian attention} where the attention logits are parameterized with a Gaussian distribution and parameters are inferred via VI and 
\item \textit{DD},  a data dependent configuration where the variational variances of the Gaussian distribution are amortized in order to support input-dependent (i.e., \emph{heteroscedastic}) uncertainties.
\end{inparaenum}

\subsection{Implicitly reparameterized Dirichlet attention}

Alternatively, we propose to directly parameterize the attention weights of each position $i$ by a Dirichlet distribution with parameter $\boldsymbol{\alpha} = a\mathbf{A_i}$, where $a$ is the sharpness parameter and $\mathbf{A_i}$ the i\textsuperscript{th} row of the scaled dot-product attention weights.
We then infer $a$ using VI.
Samples are obtained by drawing from independent Gamma distributions $X_k \sim \textrm{Gamma}(\alpha_k, 1)$ and normalizing $(\sum_{k=1}^K X_k)^{-1} \mathbf{X} \sim \textrm{Dirichlet}(\boldsymbol{\alpha})$. We further use contextual Gamma priors such that $\boldsymbol{\hat{\alpha}} \propto \mathbf{A_{i}}$, yielding an analytical KL divergence as done by \citet{joo2019dirichlet}. 
To obtain gradients of a Gamma random variable with respect to $\boldsymbol{\alpha}$, we use the implicit gradient reparametrization \citep{figurnov2019implicit}:
\begin{equation}
    \nabla_{\alpha} z = - (q_{\alpha}(z))^{-1} \nabla_{\alpha}F(z|\alpha)
\end{equation}
where $q_{\alpha}(z)$ is the Gamma density function and $F(z|\alpha)$ its CDF.
Like Gaussian attention, we consider a variation where the sharpness parameter depends on the input, referred to as data dependent.

\section{Experiments}
\vspace{-1em}

We run experiments using the transformer \cite{vaswani2017attention} and vision transformer \cite{dosovitskiy2021vit} on MNIST image classification \citep{lecun2010mnist}, Universal Dependencies part-of-speech (POS) tagging \citep{nivreuniversaldependencies} and on synthetic datasets (M1, M2).
We evaluate our models using test log-likelihood, predicted variance mean squared error and the expected mean square error on the synthetic dataset. The test log-likelihood, accuracy, F1-score, and expected calibration error (ECE) \citep{guo2017calibration} are used for experiments on the POS tagging and MNIST datasets.
We compare the results obtained by our methods with a transformer (MLE) and an ensemble of 30 transformers both trained by maximum likelihood.
Further details are given in \cref{sec:imp_details}.



\subsection{Result 1: Weight-space inference does not improve over MLE}
\label{sec:weight_vi}

\paragraph{Different posteriors do not help.}
We find that all weight-space VI methods are outperformed by both maximum-likelihood baselines with respect to all metrics and on all datasets (see \cref{table:main_results}). Interestingly, changing the posterior distribution does not significantly influence the performance, considering the large gap between the scores of the VI methods and baselines (see also~\cref{table:full_weight_inference_results} in the appendix). Furthermore, no variational posterior systematically outperforms the others.
\newline
Linearized Laplace inference (either on all parameters or just the final layer) shows much better results than VI. However, it still underperforms our baselines. Finally, even concrete dropout improves over VI and Laplace inference and is more competitive with our baselines.


\begin{table}[t]
\caption{VI and Laplace inference in weight-space compared to maximum likelihood models and concrete dropout. We see that the Bayesian transformers do not outperform the baselines.}
\centering
\resizebox{\linewidth}{!}{
\begin{tabular}{c l c c c c c c c c} 

    \textbf{Dataset} & \textbf{Metric} & \textbf{MLE} & \textbf{Ensemble} & \textbf{Gaussian VI} & \textbf{Laplace} & \textbf{Final Laplace} & \textbf{Concrete DP} & \textbf{Gauss. Attention} & \textbf{Dir. Attention} \\ 
    \hline

    \parbox[t]{2mm}{\multirow{3}{*}{\rotatebox[origin=c]{90}{M1}}} & Log-like. & -26.206 $\pm$ 0.000 & -26.011 $\pm$ 0.007 & -27.23 $\pm$ 0.01 & -26.282 $\pm$ 0.014 & -26.219 $\pm$ 0.003 & -25.767 $\pm$ 0.008 & -26.1623 $\pm$ 0.0006 & \textbf{-22.04 $\pm$ 0.01}\\ 
    & Var. MSE & 0.014 $\pm$ 0.000 & 0.0081 $\pm$ 0.0002 & 0.082 $\pm$ 0.004  & 0.021 $\pm$ 0.002 & 0.020 $\pm$ 0.003 & \textbf{0.007 $\pm$ 0.000} & 0.029 $\pm$ 0.000 & 0.430 $\pm$ 0.002 \\
    & MSE & \textbf{0.996 $\pm$ 0.000} & 1.0143 $\pm$ 0.0002 & 1.078 $\pm$ 0.001 & 1.0432 $\pm$ 0.0009 & 1.043 $\pm$ 0.002 & 1.0175 $\pm$ 0.0001 & 1.007 $\pm$ 0.000 & 1.0263 $\pm$ 0.0002 \\

    \hline
    \parbox[t]{2mm}{\multirow{3}{*}{\rotatebox[origin=c]{90}{M2}}} & Log-like. & -26.5670 $\pm$ 0.000 & -28.592 $\pm$ 0.009 & -35.43 $\pm$ 0.03 & -32.92 $\pm$ 0.05 & -32.469 $\pm$ 0.01 & -27.11 $\pm$ 0.04 & -26.374 $\pm$ 0.002 & \textbf{-24.841 $\pm$ 0.007} \\
    & Var. MSE & \textbf{16.943 $\pm$ 0.000} & 23.45 $\pm$ 0.09 & 110.57 $\pm$ 3.25 & 47.56 $\pm$ 0.06 & 47.07 $\pm$ 0.04 & 21.85 $\pm$ 0.08 & 20.9010 $\pm$ 0.0007 & 17.93 $\pm$ 0.03 \\
    & MSE & \textbf{1.170 $\pm$ 0.000} & 1.3552 $\pm$ 0.0003 & 2.95 $\pm$ 0.02 & 1.9943 $\pm$ 0.0008 & 1.972 $\pm$ 0.002 & 1.192 $\pm$ 0.001 & 1.2015 $\pm$ 0.0002 & 1.1928 $\pm$ 0.0006\\
    
    \hline
    \parbox[t]{2mm}{\multirow{4}{*}{\rotatebox[origin=c]{90}{POS}}} & Log-like. & \textbf{-3.707 $\pm$ 0.000} & -4.240 $\pm$ 0.006 & -17.86 $\pm$ 0.03 & -4.539 $\pm$ 0.000 & -4.539 $\pm$ 0.000 & -8.2004 $\pm$ 0.0001 & -3.9692 $\pm$ 0.0008 & -3.9682 $\pm$ 0.0003\\ 
    & Acc. & 0.9706 $\pm$ 0.0000 & \textbf{0.9708 $\pm$ 0.0001} & 0.871 $\pm$ 0.002 & 0.959 $\pm$ 0.000 & 0.958 $\pm$ 0.000 & 0.964 $\pm$ 0.000 & 0.969 $\pm$ 0.000 & 0.968 $\pm$ 0.000\\
    & F1 & \textbf{0.971 $\pm$ 0.000} & \textbf{0.971 $\pm$ 0.000} & 0.852 $\pm$ 0.000 & 0.959 $\pm$ 0.000 & 0.959 $\pm$ 0.000 & 0.964 $\pm$ 0.000 & 0.969 $\pm$ 0.000 & 0.968 $\pm$ 0.000\\
    & ECE & 0.03 $\pm$ 0.00 & \textbf{0.0261 $\pm$ 0.0001} & 0.052 $\pm$ 0.001 & 0.048 $\pm$ 0.000 & 0.048 $\pm$ 0.000 & 0.031 $\pm$ 0.000 & 0.0271 $\pm$ 0.0000 & 0.0287 $\pm$ 0.0000 \\

    \hline
    \parbox[t]{2mm}{\multirow{4}{*}{\rotatebox[origin=c]{90}{MNIST}}} & Log-like. & -0.074 $\pm$ 0.000 & -0.1133 $\pm$ 0.0008 & -3.18 $\pm$ 0.04 & -0.088 $\pm$ 0.000 & -0.09 $\pm$ 0.00 & \textbf{-0.064 $\pm$ 0.000} & -0.0720 $\pm$ 0.0001 & -0.1045 $\pm$ 0.0005 \\
    & Acc. & 0.979 $\pm$ 0.000 & \textbf{0.9825 $\pm$ 0.0003} & 0.101 $\pm$ 0.002 & 0.972 $\pm$ 0.000 & 0.972 $\pm$ 0.000 & 0.981 $\pm$ 0.000 & 0.9790 $\pm$ 0.0002 & 0.9738 $\pm$ 0.0003\\
    & F1 & 0.979 $\pm$ 0.000 & \textbf{0.982 $\pm$ 0.000} & 0.092 $\pm$ 0.000 & 0.972 $\pm$ 0.000 & 0.972 $\pm$ 0.000 & 0.981 $\pm$ 0.000 & 0.9786 $\pm$ 0.0000 & 0.9736 $\pm$ 0.0000\\
    & ECE & 0.022 $\pm$ 0.000 & 0.0326 $\pm$ 0.0004 & 0.097 $\pm$ 0.009 & 0.035 $\pm$ 0.000 & 0.038 $\pm$ 0.000 & \textbf{0.020 $\pm$ 0.000} & 0.0227 $\pm$ 0.0002 & 0.0305 $\pm$ 0.0003\\
\end{tabular}
}
\vspace{-1em}
\label{table:main_results}
\end{table}

\paragraph{The prior is (at least partially) at fault.}
\label{sec:weight_empirical_study}

In our attempt to understand the poor performance of weight-space VI in transformers, we conduct an empirical weight distribution study. 
We find that the marginal weight distributions are essentially uni-modal, except for some embedding and projection layers which tend to have two or three less significant modes. 
\begin{wraptable}{r}{0.5\linewidth}
\caption{Improvement of VI with improved priors relative to Gaussian priors.}
\centering
\resizebox{\linewidth}{!}{
\begin{tabular}{l c c c c c} 
\hline
    \textbf{Dataset} &  \textbf{Gauss. VI} & \textbf{Laplace VI} & \textbf{Logistic VI} & \textbf{Cauchy VI} & \textbf{Student VI} \\
    \hline
    M1 & 1.40\% & 3.80\% & 4.12\% & 1.85\% & 2.79\% \\
    M2 & 2.85\% & 3.06\% & 2.76\% & 4.36\% & 2.70\% \\
    POS & 0.12\% & 2.05\% & 2.16\% & 0.87\% & -0.32\%\\
    MNIST & 26.95\% & 33.31\% & 31.36\% & 5.66\% & 26.94\% \\
    \hline
\end{tabular}
}
\label{fig:improved_priors}
\end{wraptable} 
Furthermore, other than a decrease in tailedness (high degree of freedom) of the last layer, no recurrent pattern in the tailedness of the weight distribution appears across the considered datasets (\cref{fig:tailedness} in the appendix).
Likewise, no single distribution seems to universally fit the empirical distributions of the weights across all datasets (see \cref{fig:attention_queries_qq_plots} and \cref{fig:attention_mlp_qq_plots} in the appendix). This suggests that the shape of the weight distribution strongly depends on the considered dataset.
%
%
%
Using the observations from this weight distribution study, we choose more appropriate ("improved") priors. \Cref{fig:improved_priors} shows systematic likelihood improvements. Moreover, we find that the performance of VI critically depends on the prior parameters (see \cref{fig:ll_sensitivity_prior_scale} in the appendix).

\subsection{Results 2: Variational attention is better than weight-space inference}
\label{sec:variational_attetnion}

\paragraph{Dirichlet attention works well.}
Unlike weight-space inference, we find that inference on the attention weights works competitively. Indeed, Dirichlet attention strongly outperforms our baselines in terms of likelihood on the synthetic data and lies between both baselines on the POS tagging and MNIST (\cref{table:main_results}). However, the data dependent configuration does not systematically outperform its standard counterpart (see \cref{table:variational_attention} in the appendix). Moreover, Dirichlet attention outperforms Gaussian attention in terms of log-likelihood on the toy data and POS tagging, but not on MNIST. 


\paragraph{Variational attention leads to more consistent prior entropies.}
While investigating the entropy of the predictive distribution when sampling weights from the priors, we find that non-improved priors yield highly variable entropy distributions, ranging from low values around $1$ to higher values around $2.3$ bits. 
However, when sampling from improved priors selected by our weight distribution analysis, the entropy distribution concentrates very strongly around a high value of $2.3$ bits. This same behavior is observed when sampling from the Gaussian and our proposed Dirichlet attention.
This is desirable as the prior predictive should show high uncertainty in function space.


\begin{figure}[H]
\vspace{-1em}
\resizebox{\linewidth}{!}{
\begin{tabular}{cccccccc}
\subfloat[Gaussian]{\includegraphics[width=0.15\linewidth]{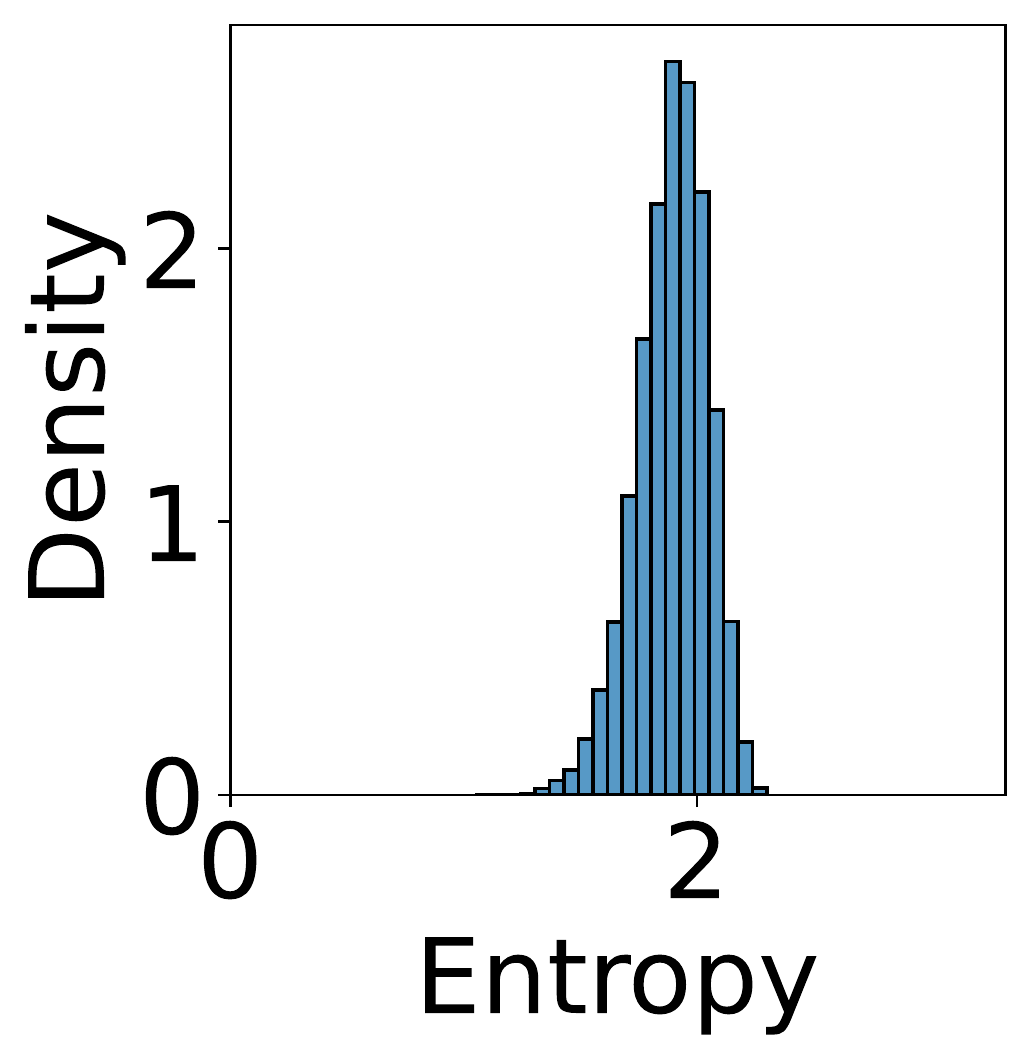}} &
\subfloat[Laplace]{\includegraphics[width=0.15\linewidth]{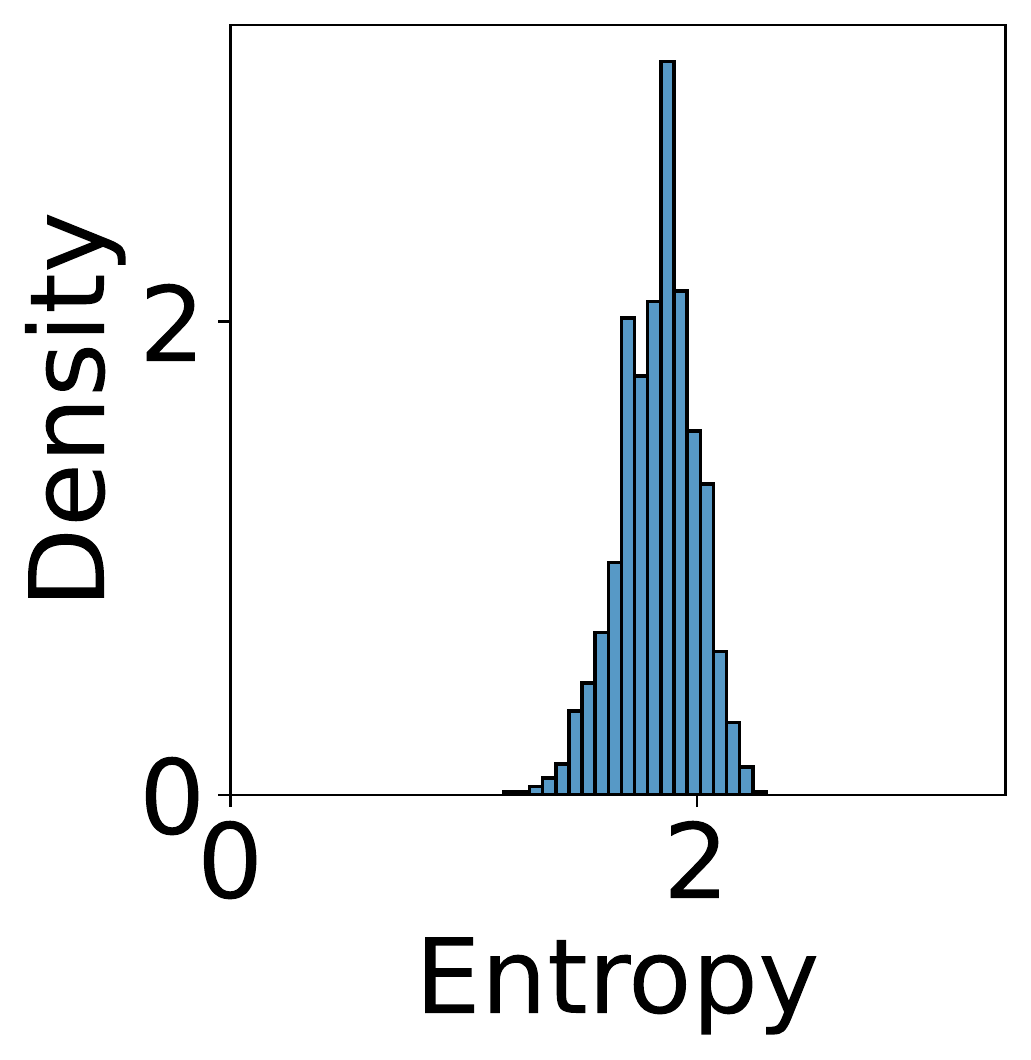}} &
\subfloat[Logistic]{\includegraphics[width=0.15\linewidth]{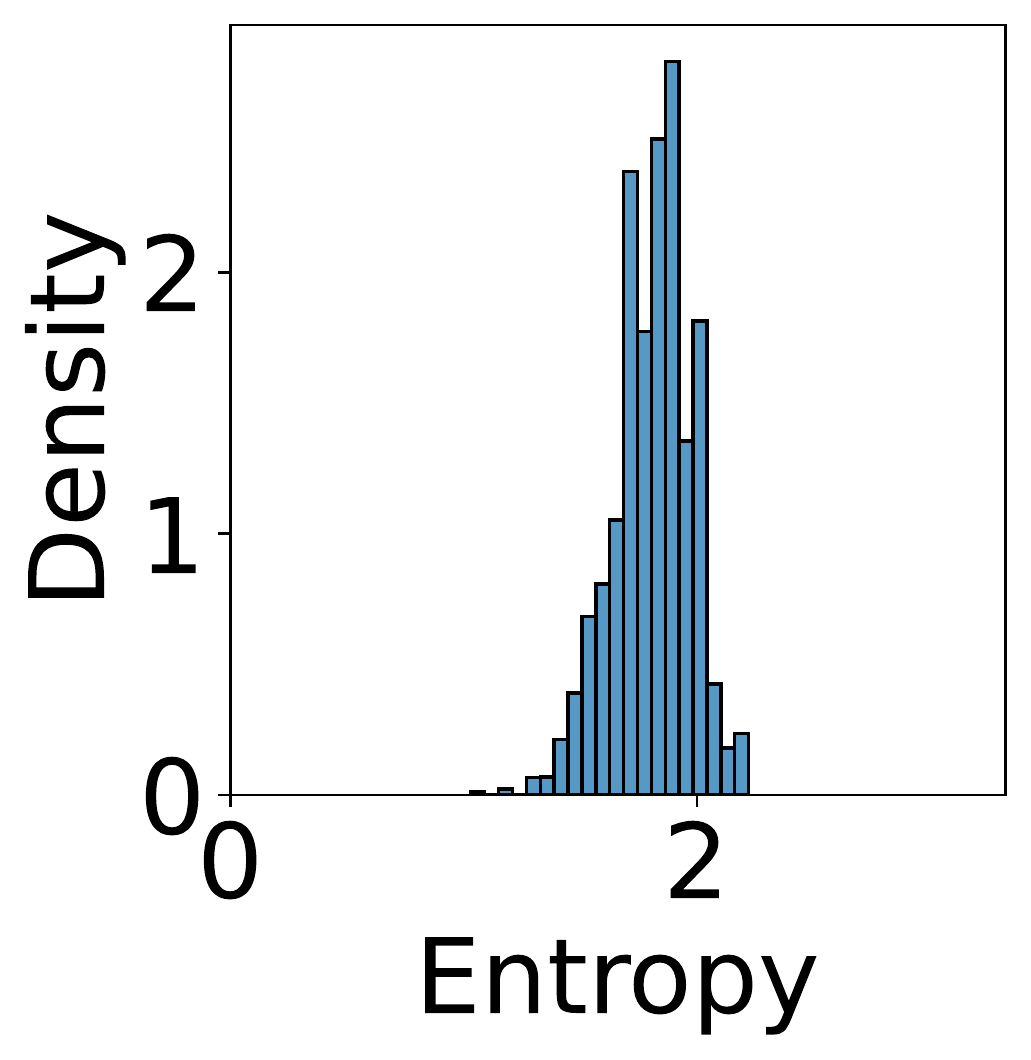}} &
\subfloat[Cauchy]{\includegraphics[width=0.15\linewidth]{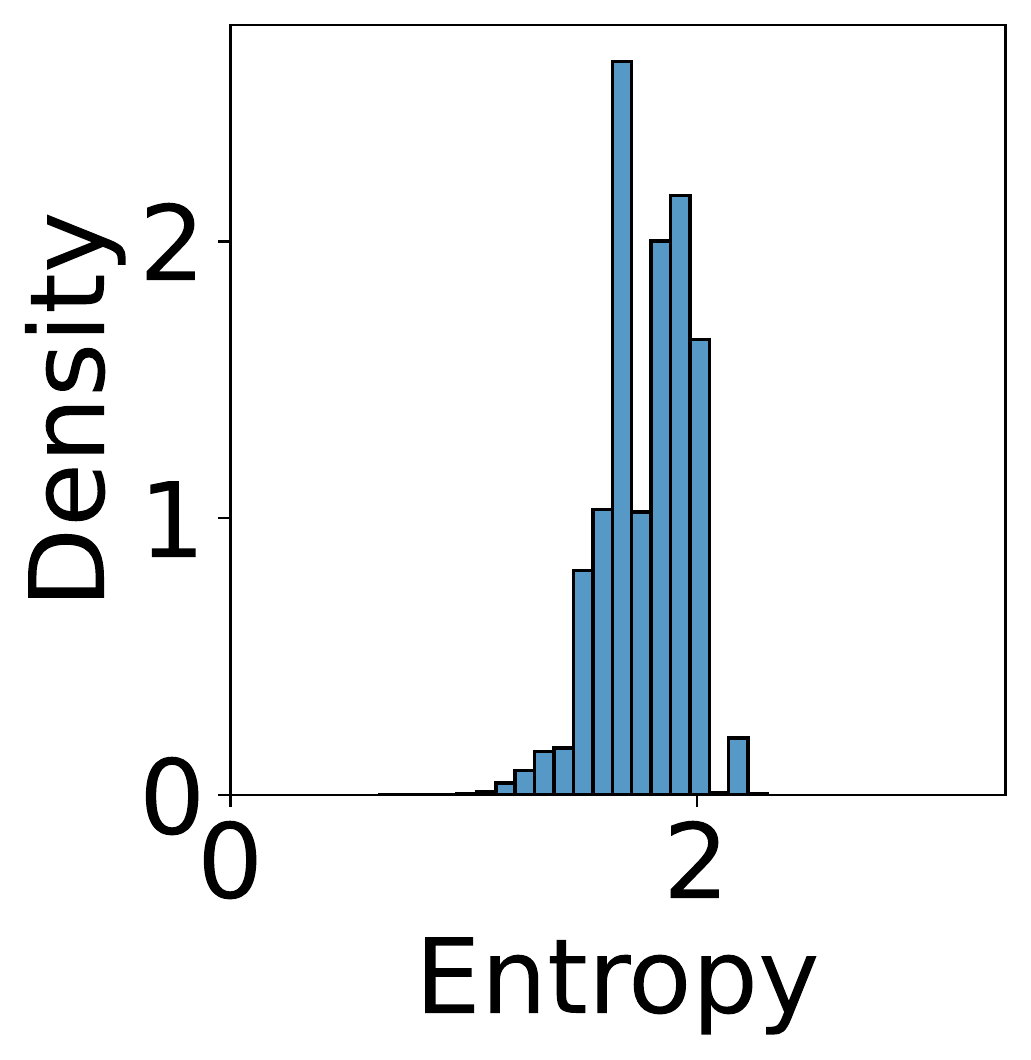}} &
\subfloat[Student]{\includegraphics[width=0.15\linewidth]{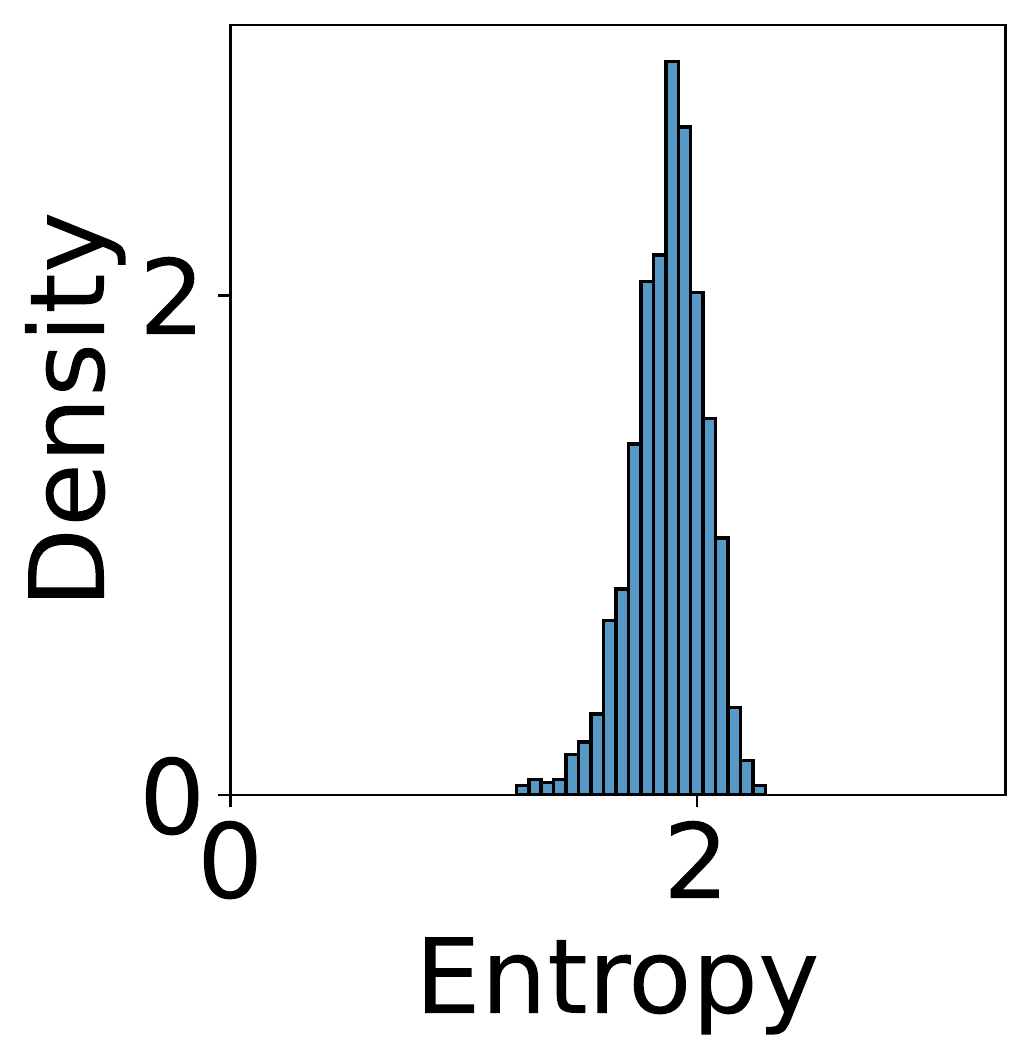}} &
\subfloat[Improved]{\includegraphics[width=0.169\linewidth]{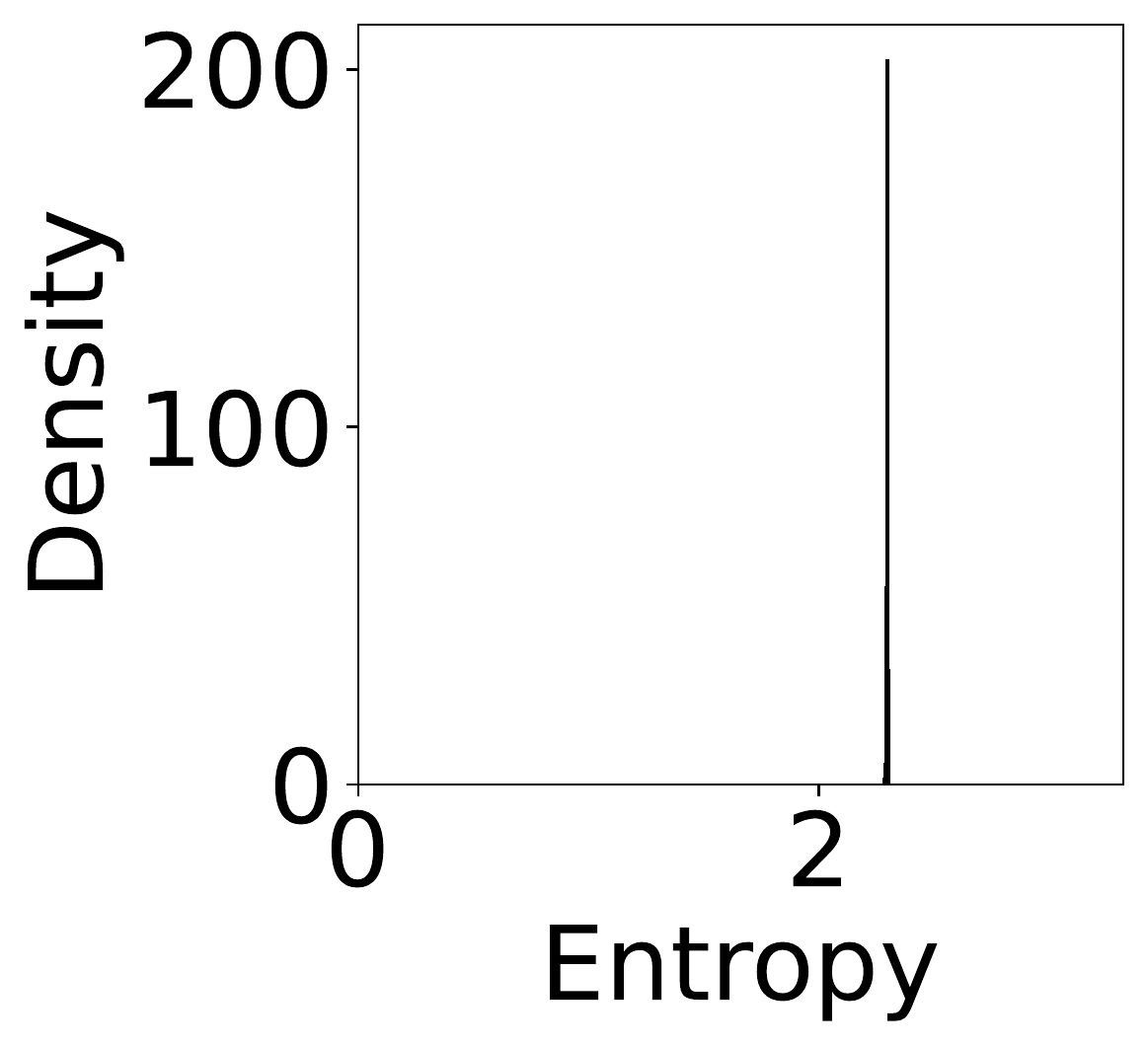}} & 
\subfloat[Gauss. att.]{\includegraphics[width=0.169\linewidth]{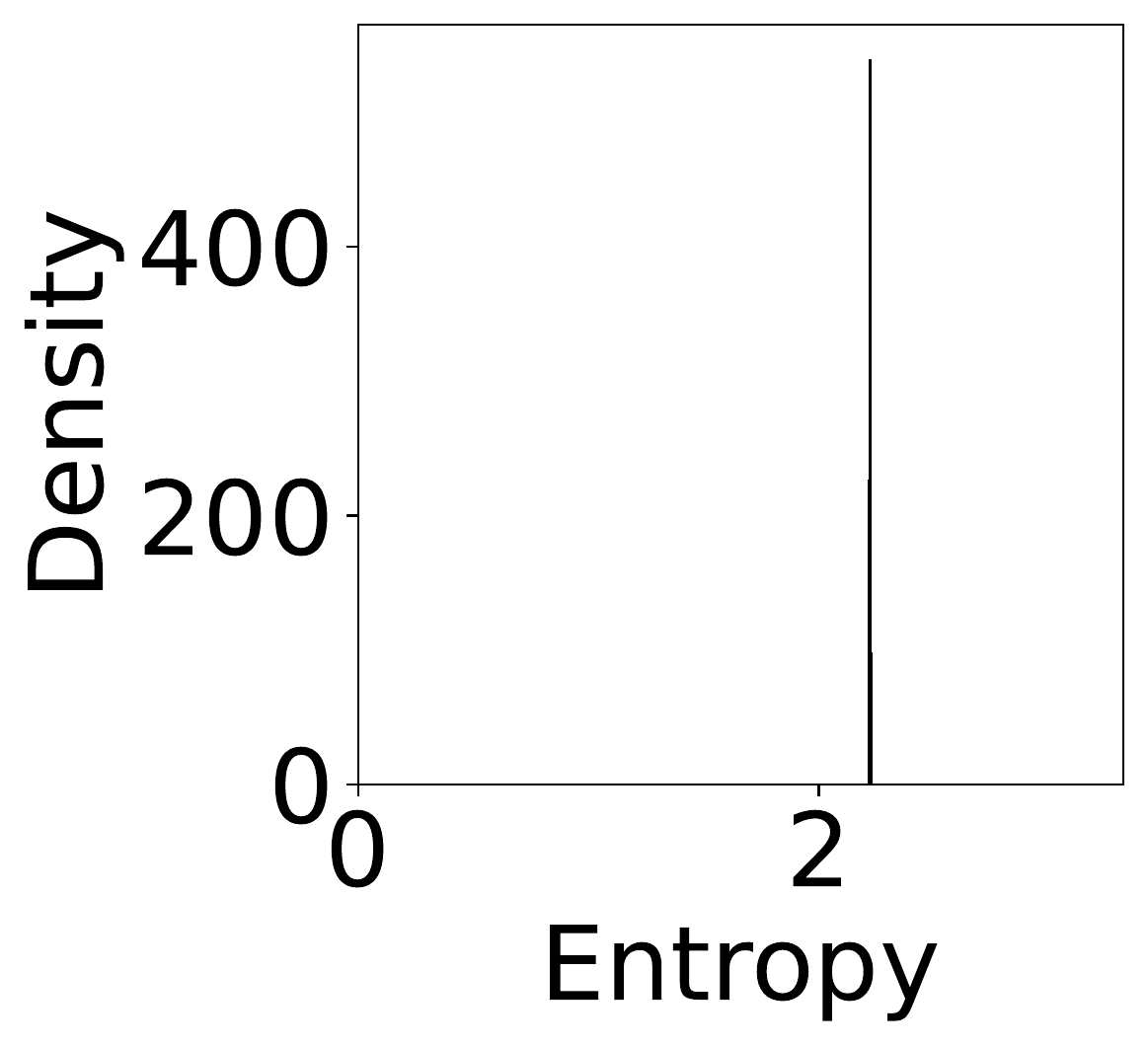}} &
\subfloat[Dir. att.]{\includegraphics[width=0.169\linewidth]{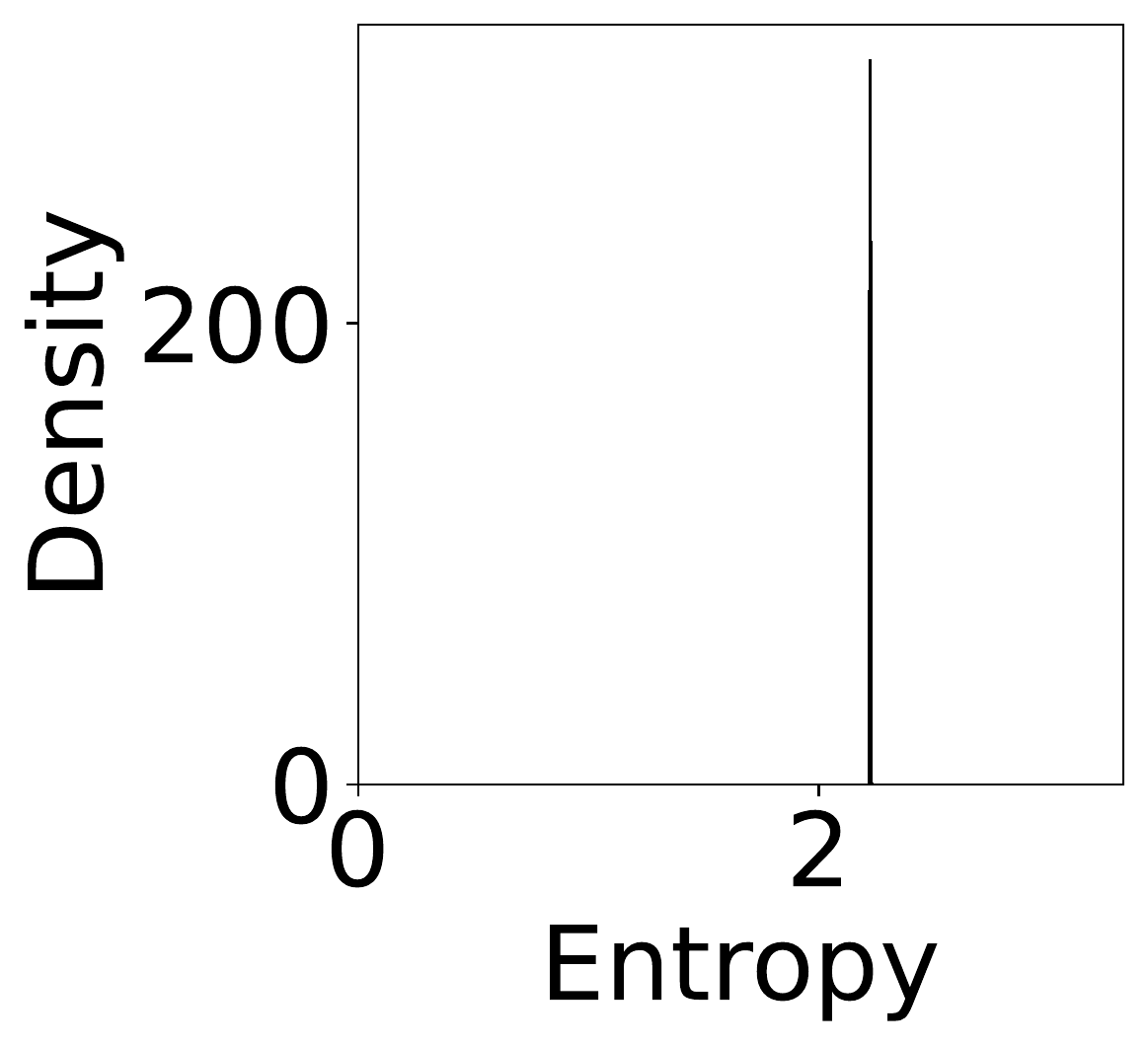}}
\end{tabular} 
}
\caption{Prior predictive entropy distributions on MNIST train data. Improving the weight-space priors and using variational attention both lead to more consistently high entropies.}
\label{fig:output_dist_entropy_sample_prior}
\end{figure}

\section{Related work}

\paragraph{Bayesian Transformers.}

Previous attempts to make the transformer Bayesian have used VI to perform inference on a subset of layers \citep{tran2019bayesian, xue2021bayesian}. While both methods claim state-of-the-art performance on their respective benchmarks, \citet{tran2019bayesian} do not provide any quantitative results and \citet{xue2021bayesian} initialize their priors to a maximum estimate of the weights which is not strictly Bayesian. Alternatively, \citet{fan2020bam} parameterize the attention logits of a transformer by a Gaussian distribution and finetune the deterministic self-attention of language models pretrained on large corpora. They however only consider finetuning and not full training using variational attention.
Orthogonally, \citet{martin2020monte} consider attention keys, queries, values, and weights as unobserved random variables and use sequential Monte Carlo methods to sample them. 

\paragraph{Bayesian neural network inference.}

BNN inference has recently advanced in terms of VI methods with more expressive posteriors \citep{dusenberry2020efficient, louizos2016maxtrixgaussianposteriors, louizos2017multiplicativenfvi, mishkin2019slang, tomczak2020lowrankgaussianvi}, more efficient inference \citep{gal2016mcdropout, gal2017concretedropout, swiatkowski2020ktied}, and greater stability \citep{kingma2015lrt, wen2018flipout}. 
Likewise, the Laplace inference for BBNs has improved in scalability using further GGN approximations \citep{immer2021scalable, immer2021glm, kristiadi2020bayesian, ritter2018scalable, ritter2018scalablelaplace} and sub-network inference \citep{daxberger2021bayesian, kristiadi2020bayesian}.
Orthogonally, MCMC methods for BNNs have been improved \citep{fortuin2021bnnpriors, garriga2021exact, wenzel2020good, zhang2019cyclical}, better BNN priors have been studied \citep{fortuin2021priors, fortuin2021bayesian}, and even deep ensembles \cite{lakshminarayanan2017simple} have been cast as approximate inference \citep{ciosek2019conservative, d2021repulsive, d2021stein, izmailov2021bayesian, pearce2018bayesian, pearce2020uncertainty, wilson2020bayesian}.

\section{Conclusion}

We have shown that weight space inference in Bayesian transformers does not work well, regardless of the choice of posterior. 
We also found that choosing priors according to an empirical weight distribution analysis improved the performance, suggesting that priors are at least partially at fault. However, we have not found the right priors to make the method competitive.
Moreover, we found evidence that na\"ive weight-space priors lead to low prior predictive entropy, and therefore do not reflect our true beliefs about the output distribution. In order to move closer to the function-space distribution, we suggested to perform inference on the attention weights rather than on parameters. We proposed a novel method based on the implicit reparameterization of the Dirichlet distribution to apply variational inference on the attention weights, which performed competitively with respect to our baselines. 

\newpage
\bibliographystyle{plainnat}


\clearpage

\appendix

\section{Appendix}

\subsection{Additional experimental results}

\subsubsection{Weight space inference}

We report in Table~\ref{table:full_weight_inference_results} the full weight space inference results table discussed in Section \ref{sec:weight_vi}. 

\begin{table}[H]
\caption{Weight space inference results in the transformer vs baselines}
\centering
\resizebox{\linewidth}{!}{
\begin{tabular}{c l c c c c c c c c c c} 

    \textbf{Dataset} & \textbf{Metric} & \textbf{MLE} & \textbf{Ensemble} & \textbf{Gaussian VI} & \textbf{Laplace VI} & \textbf{Logistic VI} & \textbf{Cauchy VI} & \textbf{Student VI} & \textbf{Concrete Dropout} & \textbf{Laplace} & \textbf{Final Laplace} \\ 
    \hline

    \parbox[t]{2mm}{\multirow{3}{*}{\rotatebox[origin=c]{90}{M1}}} & Log-likelihood & -26.2075 $\pm$ 0.0000 & -26.011 $\pm$ 0.007 & -27.23 $\pm$ 0.01 & -28.01 $\pm$ 0.16 & -28.08 $\pm$ 0.11 & -27.61 $\pm$ 0.02 & -27.68 $\pm$ 0.10 & \textbf{-25.767 $\pm$ 0.008} & -26.282 $\pm$ 0.014 & -26.219 $\pm$ 0.003 \\ 
    & Variance MSE & 0.0137 $\pm$ 0.0000 & 0.0081 $\pm$ 0.0002 & 0.082 $\pm$ 0.004 & 0.24 $\pm$ 0.05 & 0.226 $\pm$ 0.066 & 0.225 $\pm$ 0.009 & 0.161 $\pm$ 0.046 & \textbf{0.0066 $\pm$ 0.0000} & 0.021 $\pm$ 0.002 & 0.020 $\pm$ 0.003 \\
    & MSE & \textbf{0.9963 $\pm$ 0.0000} & 1.0143 $\pm$ 0.0002 & 1.078 $\pm$ 0.001 & 1.174 $\pm$ 0.016 & 1.181 $\pm$ 0.014 & 1.133 $\pm$ 0.001 & 1.128 $\pm$ 0.015 & 1.0175 $\pm$ 0.0001 & 1.0432 $\pm$ 0.0009 & 1.043 $\pm$ 0.002 \\

    \hline
    \parbox[t]{2mm}{\multirow{3}{*}{\rotatebox[origin=c]{90}{M2}}} & Log-likelihood & \textbf{-26.5670 $\pm$ 0.0000} & -28.592 $\pm$ 0.009 & -35.427 $\pm$ 0.034 & -35.86 $\pm$ 0.11 &  -35.72 $\pm$ 0.08 & -37.15 $\pm$ 0.03 & -35.43 $\pm$ 0.61 & -27.11 $\pm$ 0.04 & -32.92 $\pm$ 0.05 & -32.469 $\pm$ 0.01 \\
    & Variance MSE & \textbf{16.9430 $\pm$ 0.0000} & 23.45 $\pm$ 0.09 & 110.57 $\pm$ 3.25 & 125.08 $\pm$ 14.73 & 121.90 $\pm$ 16.25 & 130.11 $\pm$ 1.89 & 82.47 $\pm$ 15.27 & 21.85 $\pm$ 0.08 & 47.56 $\pm$ 0.06 & 47.07 $\pm$ 0.04 \\
    & MSE & \textbf{1.1700 $\pm$ 0.0000} & 1.3552 $\pm$ 0.0003 & 2.95 $\pm$ 0.02 & 3.10 $\pm$ 0.07 & 3.03 $\pm$ 0.09 & 3.45 $\pm$ 0.01 & 2.82 $\pm$ 0.09 & 1.192 $\pm$ 0.001 & 1.9943 $\pm$ 0.0008 & 1.972 $\pm$ 0.002\\
    
    \hline
    \parbox[t]{2mm}{\multirow{4}{*}{\rotatebox[origin=c]{90}{POS}}} & Log-likelihood & \textbf{-3.7066 $\pm$ 0.0000} & -4.240 $\pm$ 0.006 & -17.86 $\pm$ 0.03 & -17.69 $\pm$ 0.01 & -18.02 $\pm$ 0.04 & -17.07 $\pm$ 0.10 & -22.91 $\pm$ 0.05 & -8.2004 $\pm$ 0.0001 & -4.5388 $\pm$ 0.0000 & -4.5391 $\pm$ 0.0000 \\ 
    & Accuracy & 0.9706 $\pm$ 0.0000 & \textbf{0.9708 $\pm$ 0.0001} & 0.871 $\pm$ 0.002 & 0.8694 $\pm$ 0.0003 & 0.8689 $\pm$ 0.0004 & 0.878 $\pm$ 0.002 &  0.824 $\pm$ 0.001 & 0.9636 $\pm$ 0.0000 & 0.9585 $\pm$ 0.0000 & 0.9584 $\pm$ 0.0000\\
    & F1 & \textbf{0.9707 $\pm$ 0.0000} & 0.9706 $\pm$ 0.0000 & 0.8524 $\pm$ 0.0000 & 0.8531 $\pm$ 0.0000 & 0.8535 $\pm$ 0.0000 & 0.8594 $\pm$ 0.0000 & 0.7980 $\pm$ 0.0000 & 0.9637 $\pm$ 0.0000 & 0.9585 $\pm$ 0.0000 & 0.9585 $\pm$ 0.0000\\
    & ECE & 0.0302 $\pm$ 0.0000 & \textbf{0.0261 $\pm$ 0.0001} & 0.052 $\pm$ 0.001 & 0.0477 $\pm$ 0.0007 & 0.0498 $\pm$ 0.0007 & 0.0543 $\pm$ 0.0007 & 0.050 $\pm$ 0.001 & 0.0314 $\pm$ 0.0000 & 0.0481 $\pm$ 0.0000 & 0.0481 $\pm$ 0.0000\\

    \hline
    \parbox[t]{2mm}{\multirow{4}{*}{\rotatebox[origin=c]{90}{MNIST}}} & Log-likelihood & -0.0739 $\pm$ 0.0000 & -0.1133 $\pm$ 0.0008 & -3.179 $\pm$ 0.038 & -3.490 $\pm$ 0.125 & -3.385 $\pm$ 0.126 & -2.636 $\pm$ 0.016 & -3.183 $\pm$ 0.009 & \textbf{-0.0642 $\pm$ 0.0000} & -0.0879 $\pm$ 0.0000 & -0.0903 $\pm$ 0.0000\\
    & Accuracy & 0.9786 $\pm$ 0.0000 & \textbf{0.9825 $\pm$ 0.0003} & 0.101 $\pm$ 0.002 & 0.099 $\pm$ 0.003 & 0.099 $\pm$ 0.003 & 0.1024 $\pm$ 0.0002 & 0.099 $\pm$ 0.002 & 0.9807 $\pm$ 0.0000 & 0.9720 $\pm$ 0.0000 & 0.9720 $\pm$ 0.0000\\
    & F1 & 0.9786 $\pm$ 0.0000 & \textbf{0.9820 $\pm$ 0.0000} & 0.0923 $\pm$ 0.0000 & 0.0173 $\pm$ 0.0000 & 0.0173 $\pm$ 0.0000 & 0.0961 $\pm$ 0.0000 & 0.0173 $\pm$ 0.0000 & 0.9807 $\pm$ 0.0000 & 0.9719 $\pm$ 0.0000 & 0.9720 $\pm$ 0.0000\\
    & ECE & 0.0218 $\pm$ 0.0000 & 0.0326 $\pm$ 0.0004 & 0.097 $\pm$ 0.009 & 0.108 $\pm$ 0.010 & 0.117 $\pm$ 0.012 &  0.064 $\pm$ 0.001 & 0.110 $\pm$ 0.034 & \textbf{0.0200 $\pm$ 0.0000} & 0.0354 $\pm$ 0.0000 &  0.0377 $\pm$ 0.0000\\

\end{tabular}
}
\label{table:full_weight_inference_results}
\end{table}

\subsubsection{Sub-network variational inference}

In addition to performing inference on the entire set of model parameters presented in Section~\ref{sec:weight_vi}, we experiment sub-network variational inference. We find that this method performs better than full inference while still under-performing our baselines. Interestingly, we observe the same behavior as in full network VI, where the posterior distribution does not significantly change the result. 

\begin{table}[H]
\caption{VI on first attention layer with Gaussian priors vs baselines}
\centering
\small
\resizebox{\linewidth}{!}{
\begin{tabular}{c l c c c c c c c} 
    \textbf{Dataset} & \textbf{Metric} & \textbf{MLE} & \textbf{Ensemble} & \textbf{Gaussian VI} & \textbf{Laplace VI} & \textbf{Logistic VI} & \textbf{Cauchy VI} & \textbf{Student VI} \\
    \hline
    \parbox[t]{2mm}{\multirow{3}{*}{\rotatebox[origin=c]{90}{M1}}} 
    & Log-likelihood & -26.2075 $\pm$ 0.0000 & \textbf{-26.0107 $\pm$ 0.0067} & -26.1726 $\pm$ 0.0013 & -26.1723 $\pm$ 0.0002 & -26.1724 $\pm$ 0.0001 & -26.1802 $\pm$ 0.0055 & -26.1844 $\pm$ 0.0007\\ 
    & Variance MSE & 0.0137 $\pm$ 0.0000 & 0.0081 $\pm$ 0.0002 & 0.0083 $\pm$ 0.0000 & 0.0081 $\pm$ 0.0003 & 0.0082 $\pm$ 0.0003 & \textbf{0.0067 $\pm$ 0.0000} & 0.0081 $\pm$ 0.0001\\
    & MSE & \textbf{0.9963 $\pm$ 0.0000} & 1.0143 $\pm$ 0.0002 & 1.0006 $\pm$ 0.0000 & 1.0006 $\pm$ 0.0007 & 1.0006 $\pm$ 0.0006 & 1.0026 $\pm$ 0.0001 & 1.0017 $\pm$ 0.0003\\
    \hline
    
    \parbox[t]{2mm}{\multirow{3}{*}{\rotatebox[origin=c]{90}{M2}}} 
    & Log-likelihood & \textbf{-26.5670 $\pm$ 0.0000} & -28.5916 $\pm$ 0.0085 & -30.5040 $\pm$ 0.0039 & -30.5249 $\pm$ 0.0059 & -30.5258 $\pm$ 0.0069 & -30.7045 $\pm$ 0.0080 & -30.6088 $\pm$ 0.0055 \\ 
    & Variance MSE & \textbf{16.9430 $\pm$ 0.0000} & 23.4497 $\pm$ 0.0921 & 61.2115 $\pm$ 0.1124 & 61.4751 $\pm$ 0.5851 & 61.4699 $\pm$ 0.5920 & 61.3169 $\pm$ 0.1897 & 58.3248 $\pm$ 0.2691 \\
    & MSE & \textbf{1.1700 $\pm$ 0.0000} & 1.3552 $\pm$ 0.0003 & 1.8599 $\pm$ 0.0005 & 1.8637 $\pm$ 0.0054 & 1.8636 $\pm$ 0.0051 & 1.8992 $\pm$ 0.0012 & 1.8695 $\pm$ 0.0005 \\
    \hline
    
    \parbox[t]{2mm}{\multirow{3}{*}{\rotatebox[origin=c]{90}{POS}}}
    & Log-likelihood & \textbf{-3.7066 $\pm$ 0.0000} & -4.2401 $\pm$ 0.0059 & -4.3838 $\pm$ 0.0003 & -4.3836 $\pm$ 0.0017 & -4.3835 $\pm$ 0.0016 & -4.3857 $\pm$ 0.0002 & -4.4075 $\pm$ 0.0013 \\ 
    & Accuracy & 0.9706 $\pm$ 0.0000 & \textbf{0.9708 $\pm$ 0.0001} & 0.9589 $\pm$ 0.0000 & 0.9589 $\pm$ 0.0000 & 0.9589 $\pm$ 0.0000 & 0.9588 $\pm$ 0.0001 & 0.9581 $\pm$ 0.0000\\
    & F1 & \textbf{0.9707 $\pm$ 0.0000} & 0.9706 $\pm$ 0.0000 & 0.9588 $\pm$ 0.0000 & 0.9589 $\pm$ 0.0000 & 0.9589 $\pm$ 0.0000 & 0.9588 $\pm$ 0.0000 & 0.9579 $\pm$ 0.0000\\
    & ECE & 0.0302 $\pm$ 0.0000 & \textbf{0.0261 $\pm$ 0.0001} & 0.0336 $\pm$ 0.0000 & 0.0335 $\pm$ 0.0001 & 0.0335 $\pm$ 0.0001 & 0.0415 $\pm$ 0.0000 & 0.0333 $\pm$ 0.0000\\
    \hline

    \parbox[t]{2mm}{\multirow{3}{*}{\rotatebox[origin=c]{90}{MNIST}}}
    & Log-likelihood & \textbf{-0.0739 $\pm$ 0.0000} & -0.1133 $\pm$ 0.0008 & -0.1352 $\pm$ 0.0001 & -0.1298 $\pm$ 0.0001 & -0.1333 $\pm$ 0.0003 & -0.1313 $\pm$ 0.0002 & -0.1370 $\pm$ 0.0004\\
    & Accuracy & 0.9786 $\pm$ 0.0000 & \textbf{0.9825 $\pm$ 0.0003} & 0.9592 $\pm$ 0.0001 & 0.9615 $\pm$ 0.0002 & 0.9614 $\pm$ 0.0001 & 0.9594 $\pm$ 0.0002 & 0.9583 $\pm$ 0.0000\\
    & F1 & 0.9786 $\pm$ 0.0000 & \textbf{0.9820 $\pm$ 0.0000} & 0.9595 $\pm$ 0.0000 & 0.9616 $\pm$ 0.0000 & 0.9615 $\pm$ 0.0000 & 0.9592 $\pm$ 0.0000 & 0.9584 $\pm$ 0.0000\\
    & ECE & \textbf{0.0218 $\pm$ 0.0000} & 0.0326 $\pm$ 0.0004 & 0.0411 $\pm$ 0.0003 & 0.0395 $\pm$ 0.0002 & 0.0404 $\pm$ 0.0001 & 0.0276 $\pm$ 0.0003 & 0.0424 $\pm$ 0.0001\\
\end{tabular}
}
\end{table}

\subsubsection{Variational attention}

We report in Table~\ref{table:variational_attention} the full variational attention results table discussed in Section \ref{sec:variational_attetnion} including the data dependent configurations. 

\begin{table}[H]
\caption{Variational attention methods vs baselines}
\centering
\resizebox{\linewidth}{!}{
\begin{tabular}{c l c c c c c c} 

    \textbf{Dataset} & \textbf{Metric} & \textbf{MLE} & \textbf{Ensemble} & \textbf{Gauss. Attention} & \textbf{Gauss. DD Attention} & \textbf{Dir. Attention} & \textbf{Dir. DD Attention} \\ 
    \hline
    
    \multirow{3}{*}{M1} & Log-likelihood & -26.208 $\pm$ 0.000 & -26.011 $\pm$ 0.007 & -26.1623 $\pm$ 0.0006 & -26.1799 $\pm$ 0.0001 & \textbf{-22.04 $\pm$ 0.01} & -25.242 $\pm$ 0.006\\
    & Variance MSE & 0.014 $\pm$ 0.000 & \textbf{0.0081 $\pm$ 0.0002} & 0.029 $\pm$ 0.000 & 0.045 $\pm$ 0.000 & 0.430 $\pm$ 0.002 & 0.1012 $\pm$ 0.0004\\
    & MSE & \textbf{0.996 $\pm$ 0.000} & 1.0143 $\pm$ 0.0002 & 1.007 $\pm$ 0.000 & 1.007 $\pm$ 0.000 & 1.0263 $\pm$ 0.0002 & 1.0417 $\pm$ 0.0003\\

    \hline
    \multirow{3}{*}{M2} & Log-likelihood & -26.567 $\pm$ 0.000 & -28.592 $\pm$ 0.009 & -26.374 $\pm$ 0.002 & -25.3282 $\pm$ 0.0003 & \textbf{-24.841 $\pm$ 0.007} & -26.263 $\pm$ 0.004\\
    & Variance MSE & \textbf{16.943 $\pm$ 0.000} & 23.45 $\pm$ 0.09 & 20.9010 $\pm$ 0.0007 & 18.528 $\pm$ 0.002 & 17.93 $\pm$ 0.03 & 20.17 $\pm$ 0.02\\
    & MSE & 1.17 $\pm$ 0.00 & 1.3552 $\pm$ 0.0003 & 1.2015 $\pm$ 0.0002 & \textbf{1.089 $\pm$ 0.000} & 1.1928 $\pm$ 0.0006 & 1.3018 $\pm$ 0.0001\\
    
    \hline
    \multirow{4}{*}{POS} & Log-likelihood & \textbf{-3.707 $\pm$ 0.000} & -4.240 $\pm$ 0.006 & -3.9692 $\pm$ 0.0008 & -4.0934 $\pm$ 0.0005 & -3.9682 $\pm$ 0.0003 & -3.859 $\pm$ 0.002\\ 
    & Accuracy & 0.9706 $\pm$ 0.0000 & \textbf{0.9708 $\pm$ 0.0001} & 0.969 $\pm$ 0.000 & 0.969 $\pm$ 0.000 & 0.968 $\pm$ 0.000 & 0.969 $\pm$ 0.000\\
    & F1 & \textbf{0.9707 $\pm$ 0.0000} & 0.9706 $\pm$ 0.0000 & 0.969 $\pm$ 0.000 & 0.969 $\pm$ 0.000 & 0.968 $\pm$ 0.000 & 0.969 $\pm$ 0.000\\
    & ECE & 0.0302 $\pm$ 0.0000 & \textbf{0.0261 $\pm$ 0.0001} & 0.0271 $\pm$ 0.0000 & 0.0270 $\pm$ 0.0000 & 0.0287 $\pm$ 0.0000 & 0.0278 $\pm$ 0.0001 \\
    
    \hline
    \multirow{4}{*}{MNIST} & Log-likelihood & -0.0739 $\pm$ 0.0000 & -0.1133 $\pm$ 0.0008 & \textbf{-0.0720 $\pm$ 0.0001} & -0.0838 $\pm$ 0.0001 & -0.1045 $\pm$ 0.0005 & -0.0955 $\pm$ 0.0009\\ 
    & Accuracy & 0.9786 $\pm$ 0.0000 & \textbf{0.9825 $\pm$ 0.0003} & 0.9790 $\pm$ 0.0002 & 0.9769 $\pm$ 0.0001 & 0.9738 $\pm$ 0.0003 & 0.9766 $\pm$ 0.0002\\
    & F1 & 0.9786 $\pm$ 0.0000 & \textbf{0.9820 $\pm$ 0.0000} & 0.9786 $\pm$ 0.0000 & 0.9769 $\pm$ 0.0000 & 0.9736 $\pm$ 0.0000 & 0.9764 $\pm$ 0.0000 \\
    & ECE & \textbf{0.0218 $\pm$ 0.0000} & 0.0326 $\pm$ 0.0004 & 0.0227 $\pm$ 0.0002 & 0.0252 $\pm$ 0.0001 & 0.0305 $\pm$ 0.0003 & 0.0281 $\pm$ 0.0000 \\
\end{tabular}
}
\label{table:variational_attention}
\end{table}

\subsubsection{Likelihood sensitivity to the prior}

\begin{figure}[H]
\centering
\resizebox{\linewidth}{!}{
\begin{tabular}{cc}
\subfloat[M1]{\includegraphics[width=0.5\linewidth]{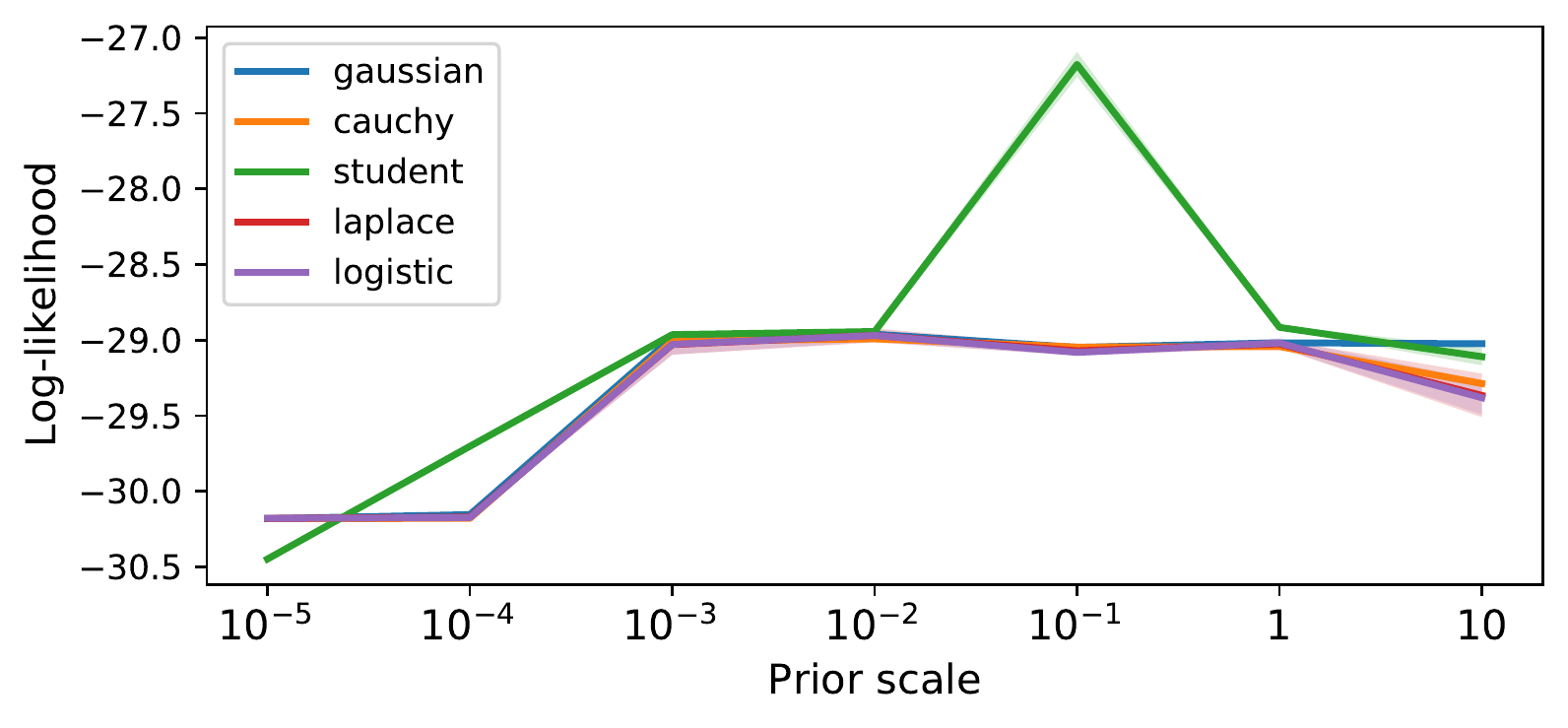}} &
\subfloat[M2]{\includegraphics[width=0.5\linewidth]{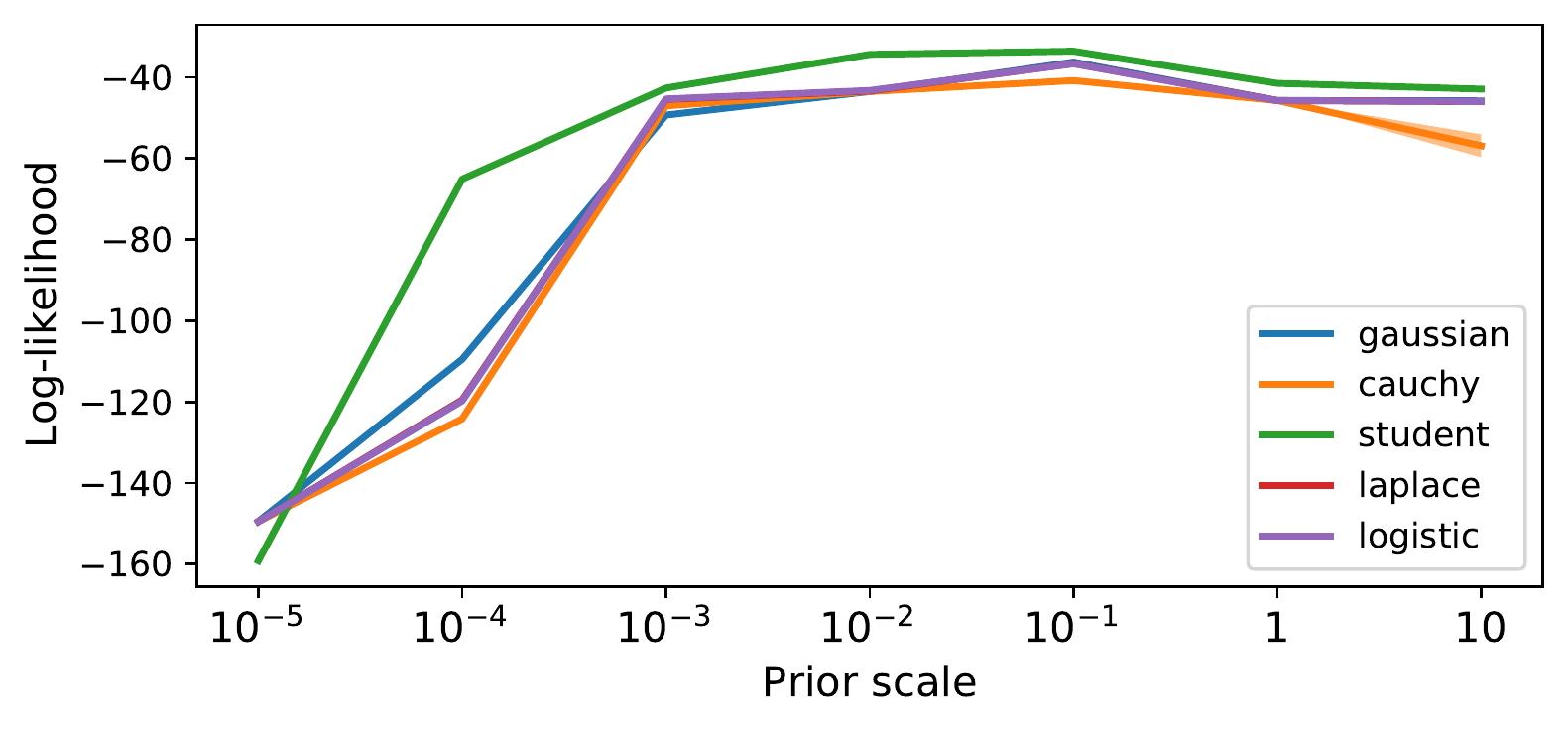}}
\end{tabular}
}
\caption{Log-likelihood sensitivity to the choice of prior scale}
\label{fig:ll_sensitivity_prior_scale}
\end{figure}

As discussed in Section~\ref{sec:weight_empirical_study}, we find that the model test likelihood is very sensitive to the choice of prior (see Figure~\ref{fig:ll_sensitivity_prior_scale}). 

\subsubsection{Weight distribution tailedness}

\begin{figure}[H]
\resizebox{\linewidth}{!}{
\begin{tabular}{cc}
\subfloat[M1, M2 \& POS tagging]{\includegraphics[width=0.5\linewidth]{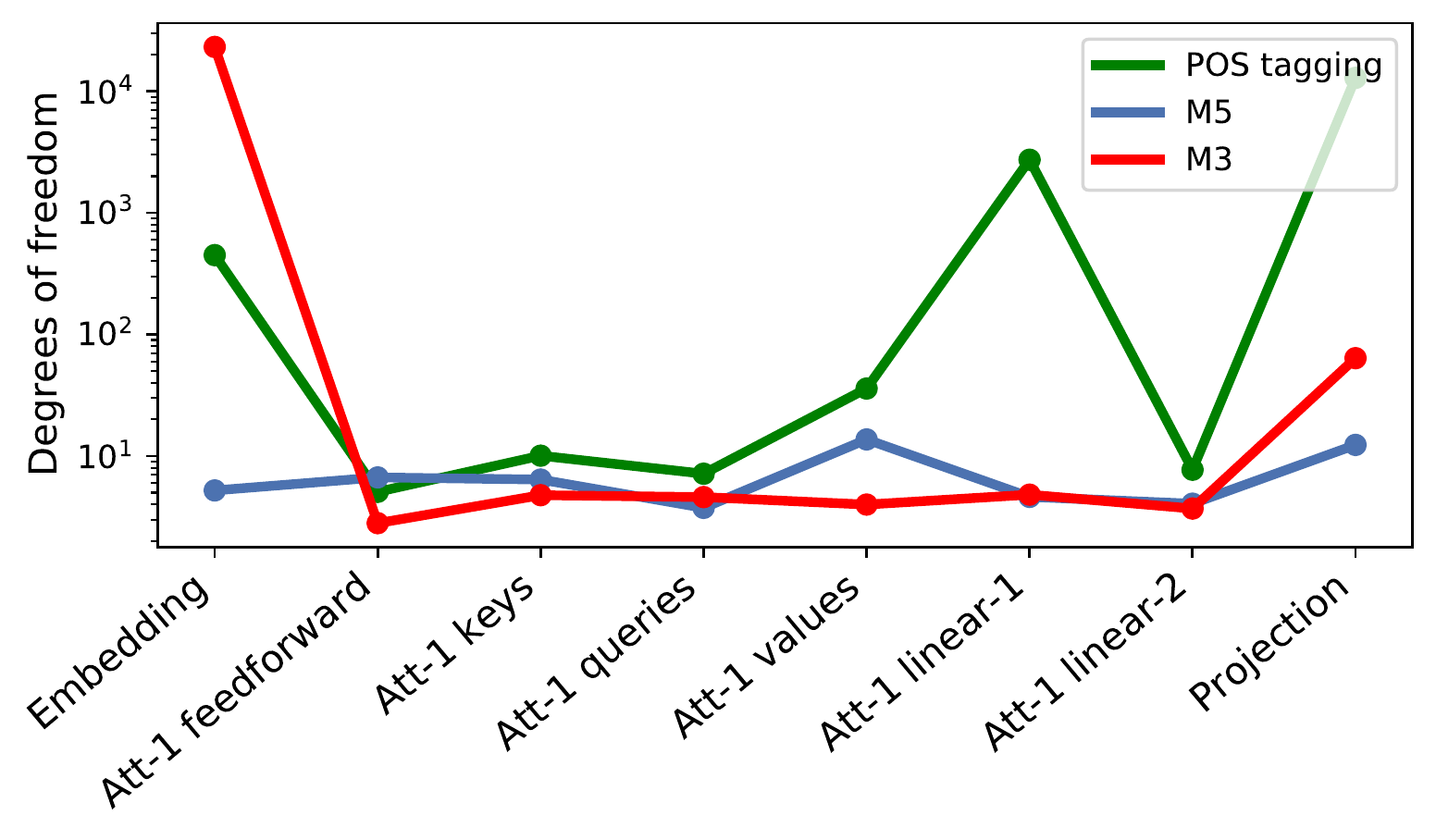}} & 
\subfloat[MNIST]{\includegraphics[width=0.5\linewidth]{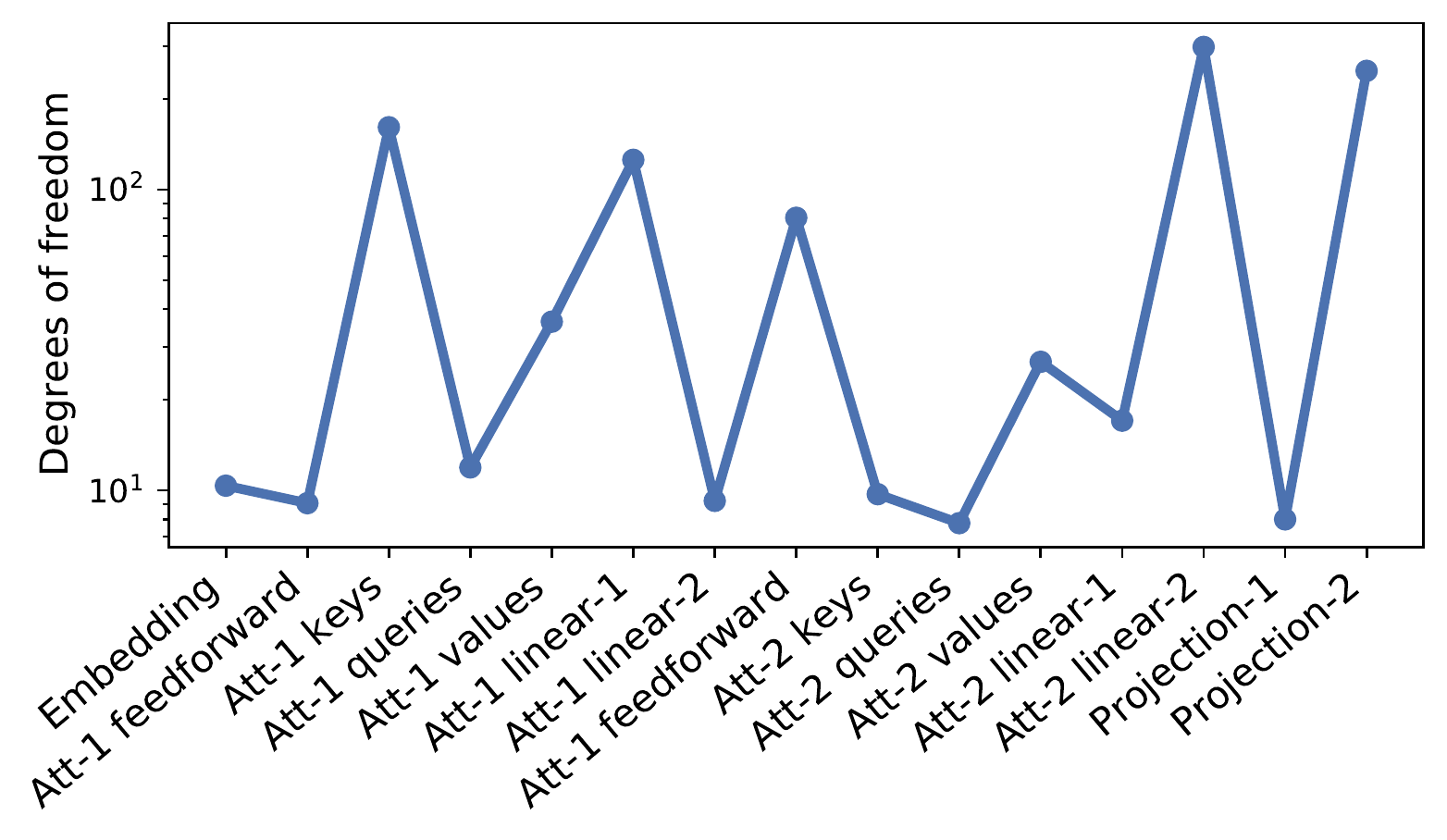}}
\end{tabular}
}
\caption{Marginal weight distribution tailedness. No patterns in the thickness of the tails, except for a decrease at the last layer, appears across the considered tasks.}
\label{fig:tailedness}
\end{figure}

\begin{figure}[H]
\resizebox{\linewidth}{!}{
\begin{tabular}{cc}
\subfloat[M1]{\includegraphics[width=0.8\linewidth, trim=0 0 110 0, clip]{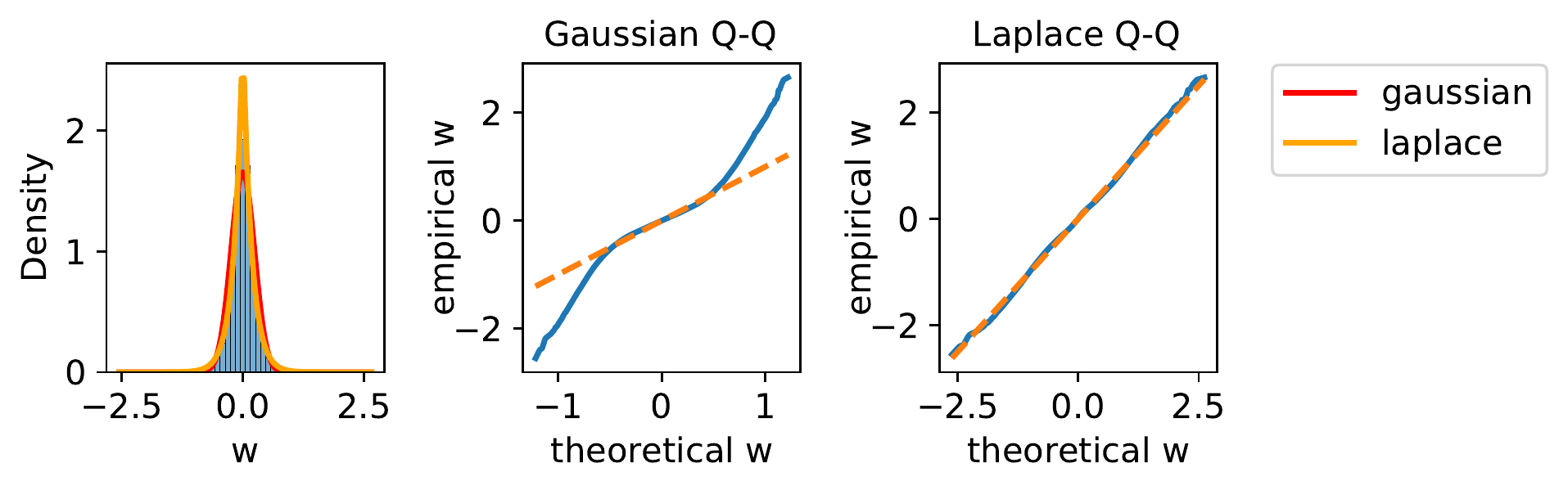}} &
\subfloat[M2]{\includegraphics[width=\linewidth]{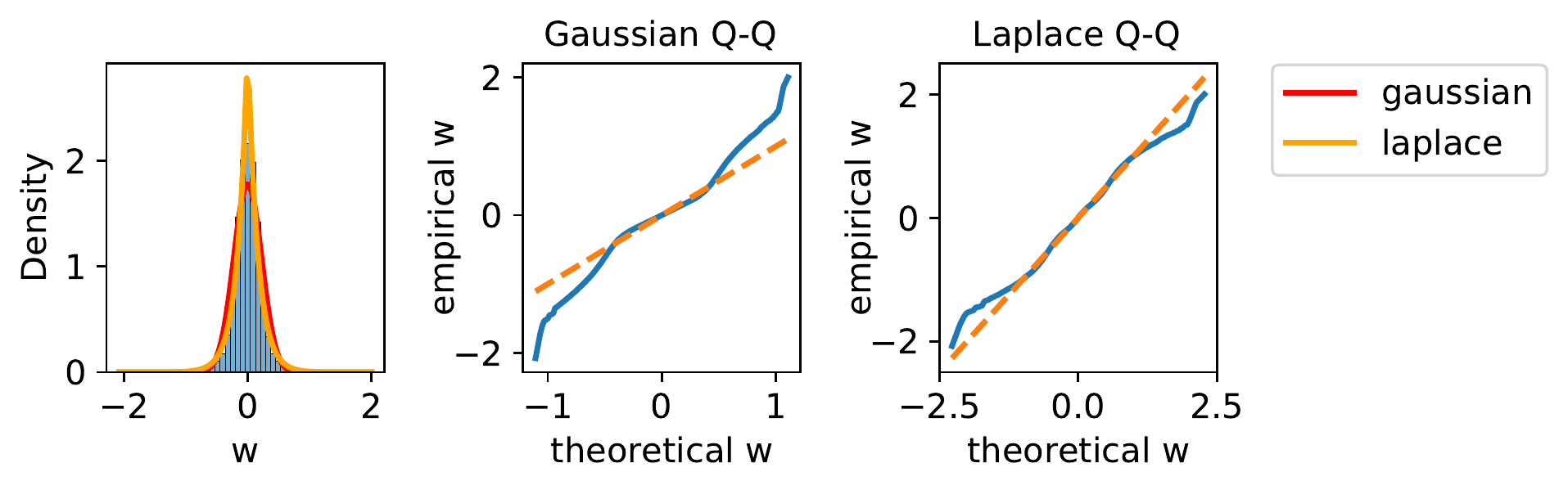}} \\
\subfloat[POS Tagging]{\includegraphics[width=0.8\linewidth, trim=0 0 110 0, clip]{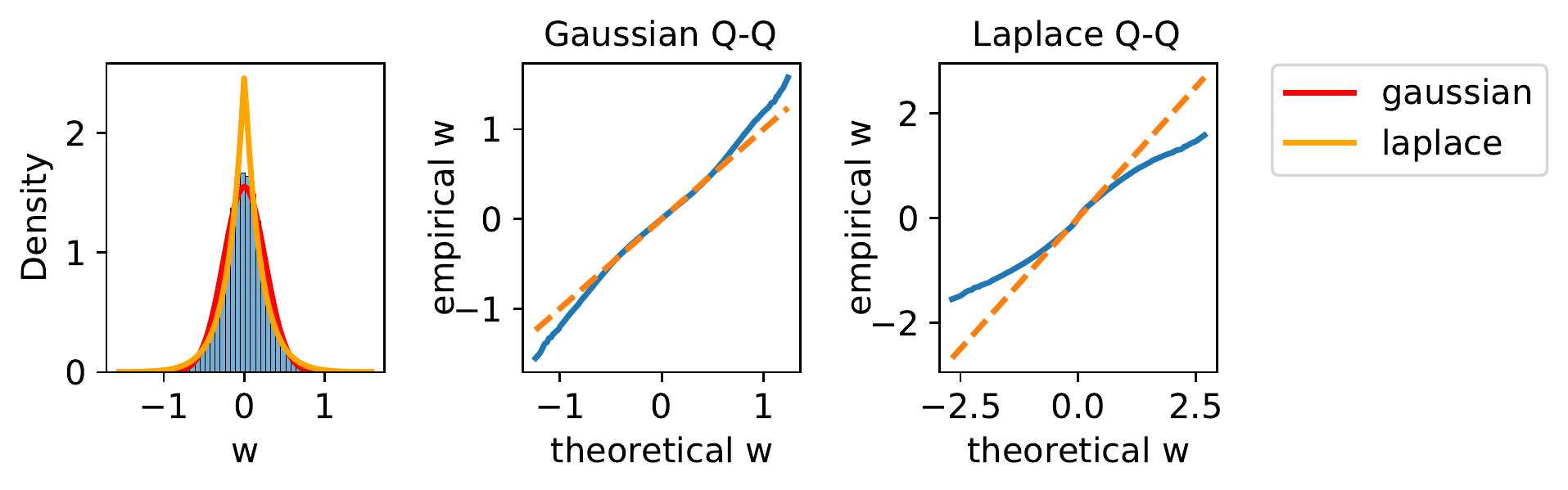}} &
\subfloat[MNIST]{\includegraphics[width=\linewidth]{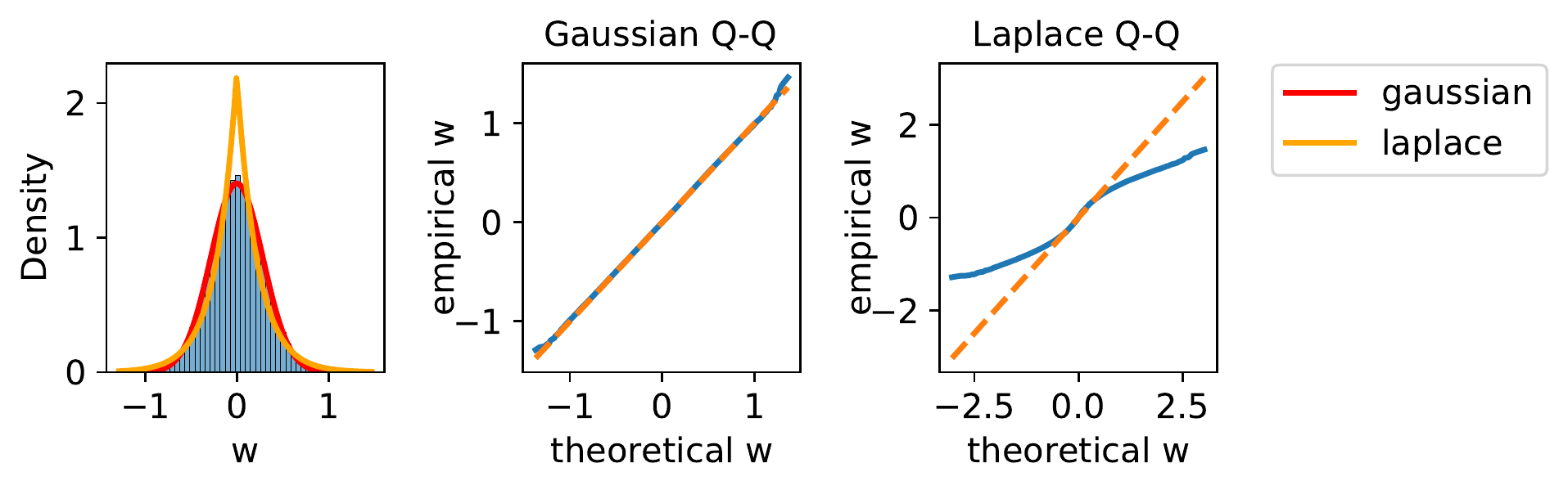}}
\end{tabular}
}
\caption{Attention queries matrix weight distribution Q-Q plot.}
\label{fig:attention_queries_qq_plots}
\end{figure}

\begin{figure}[H]
\resizebox{\linewidth}{!}{
\begin{tabular}{cc}
\subfloat[M1]{\includegraphics[width=0.8\linewidth, trim=0 0 110 0, clip]{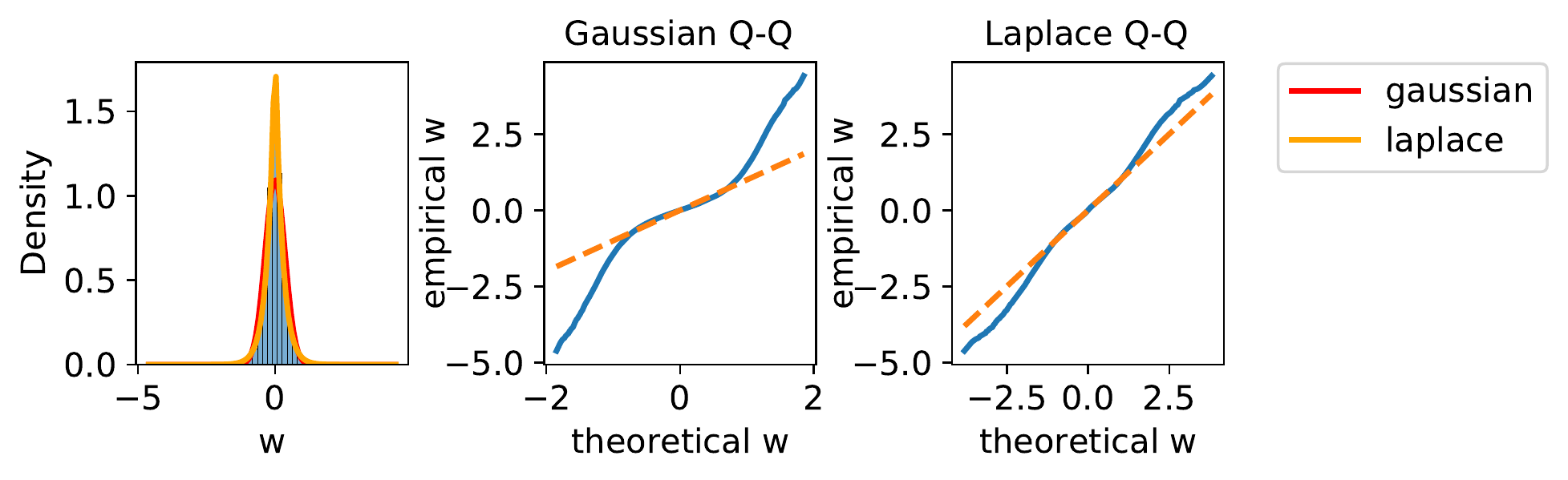}} &
\subfloat[M2]{\includegraphics[width=\linewidth]{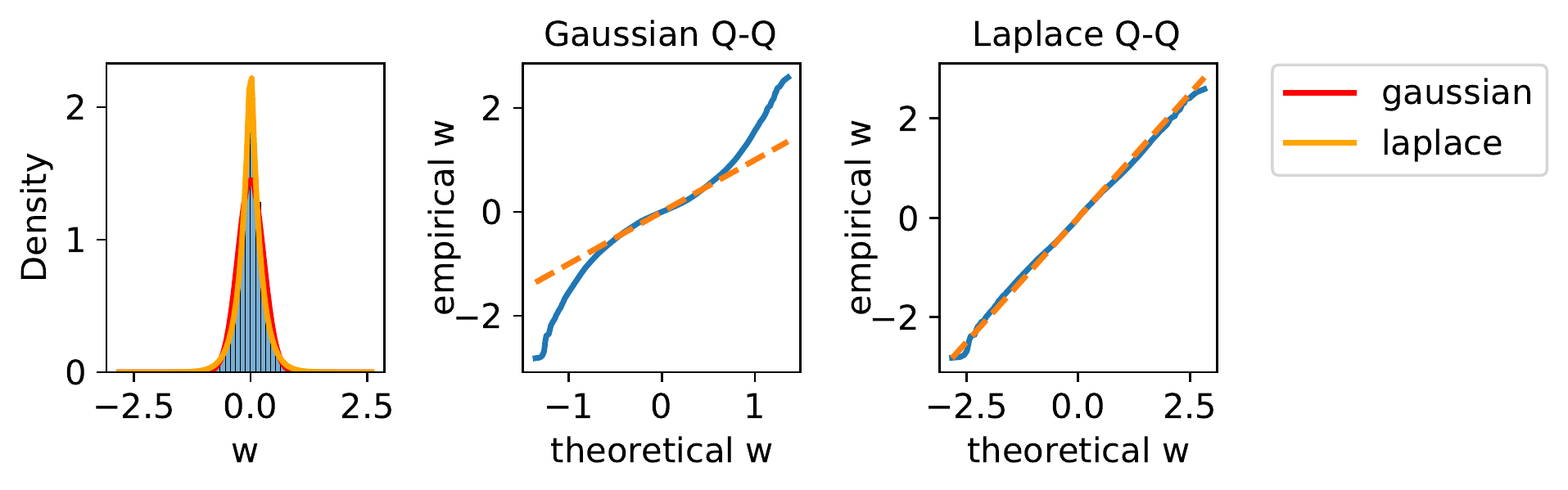}} \\
\subfloat[POS Tagging]{\includegraphics[width=0.8\linewidth, trim=0 0 110 0, clip]{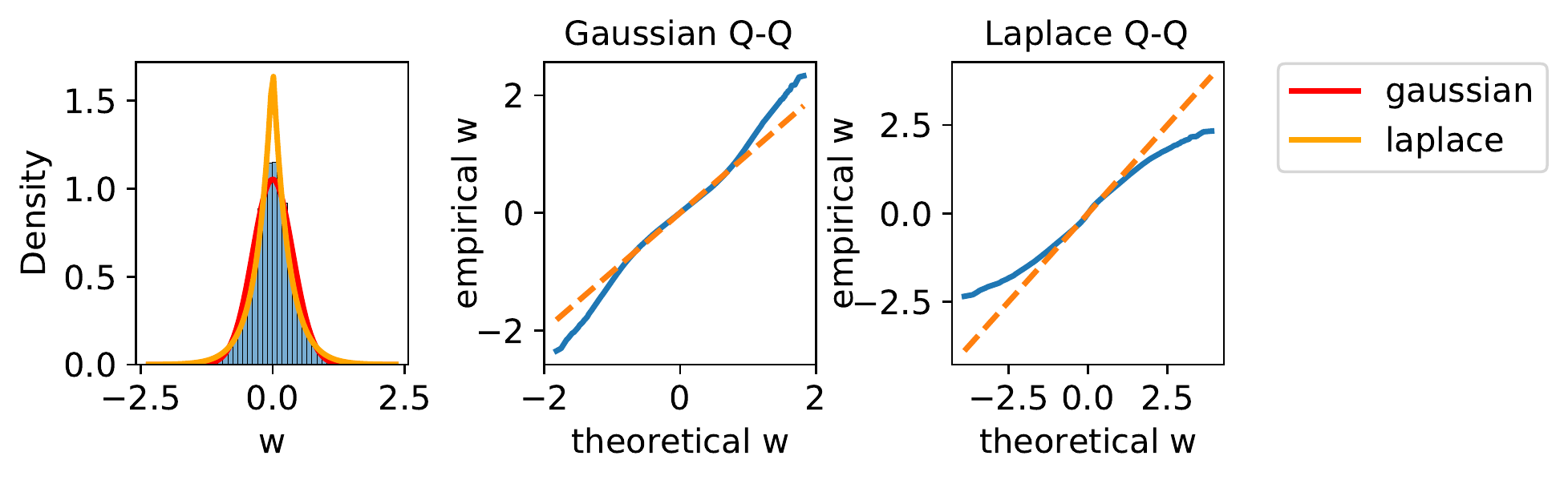}} &
\subfloat[MNIST]{\includegraphics[width=\linewidth]{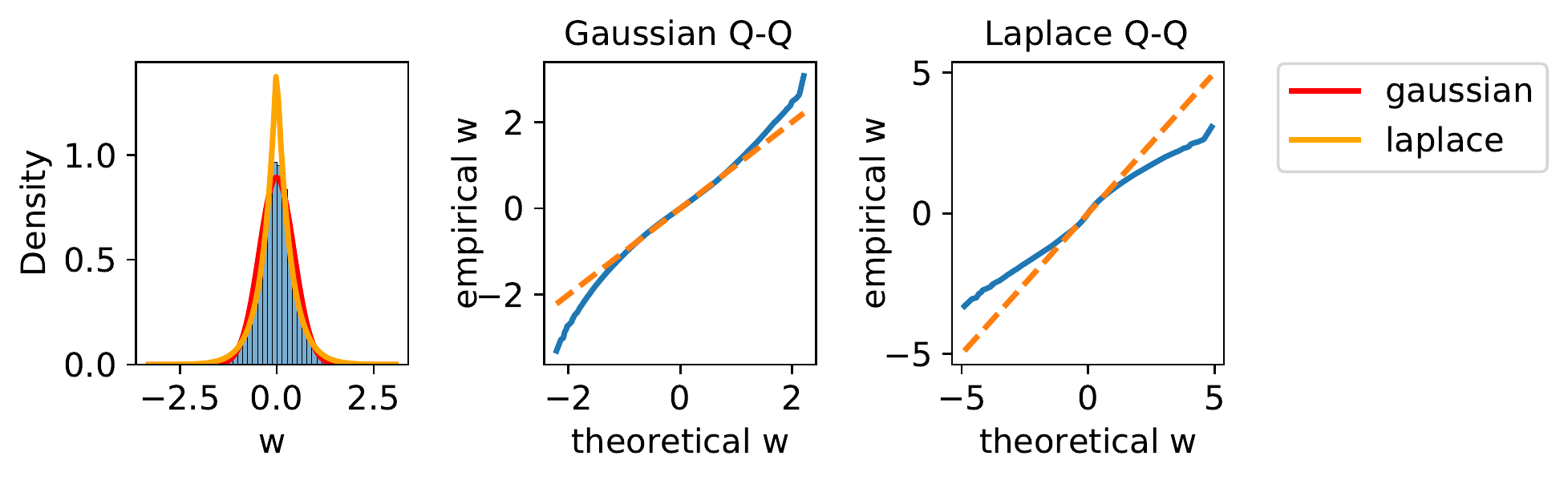}}
\end{tabular}
}
\caption{Attention MLP hidden layer empirical weight distribution Q-Q plot.}
\label{fig:attention_mlp_qq_plots}
\end{figure}

We here present more detailed results regarding the tailedness of the weight distribution. We find that no obvious patterns in the thickness of tails exist across the considered dataset (Figure~\ref{fig:tailedness}). Q-Q plots of the empirical weight distribution against light-tailed Gaussian and heavier tailed Laplace distribution provide more evidence of this phenomenon. Indeed, the weight distributions of transformers trained on MNIST and POS tagging are well fitted by a Gaussian, while transformers trained on M1 and M2 toy datasets are well fitted by a Laplace (Figure \ref{fig:attention_queries_qq_plots} and \ref{fig:attention_mlp_qq_plots}). This suggest that not one universal distribution fits the empirical distribution of the weights across all datasets.

\subsubsection{Off-diagonal covariance value distributions}

\begin{figure}[H]
\centering
\resizebox{\linewidth}{!}{
\begin{tabular}{ccc}
\subfloat[Attention queries]{\includegraphics[width=0.3\linewidth]{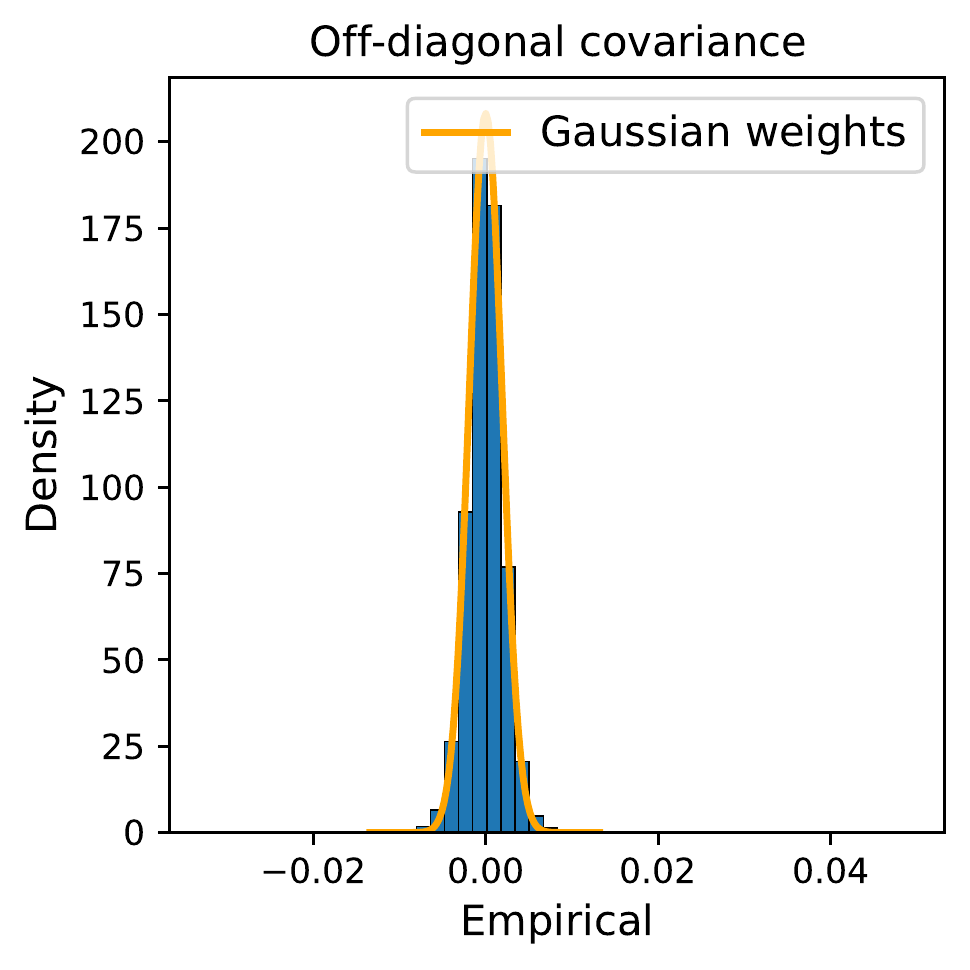}} &
\subfloat[Attention feedforward]{\includegraphics[width=0.3\linewidth]{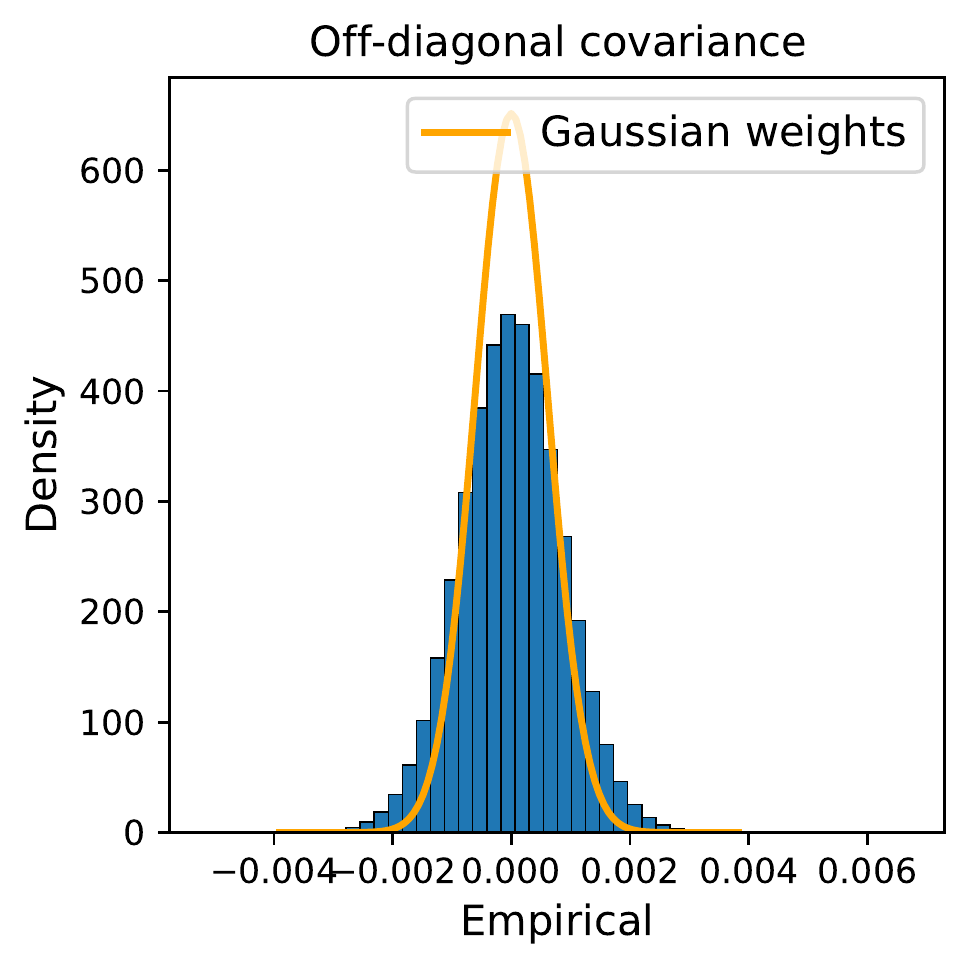}} &
\subfloat[Attention MLP hidden layer]{\includegraphics[width=0.3\linewidth]{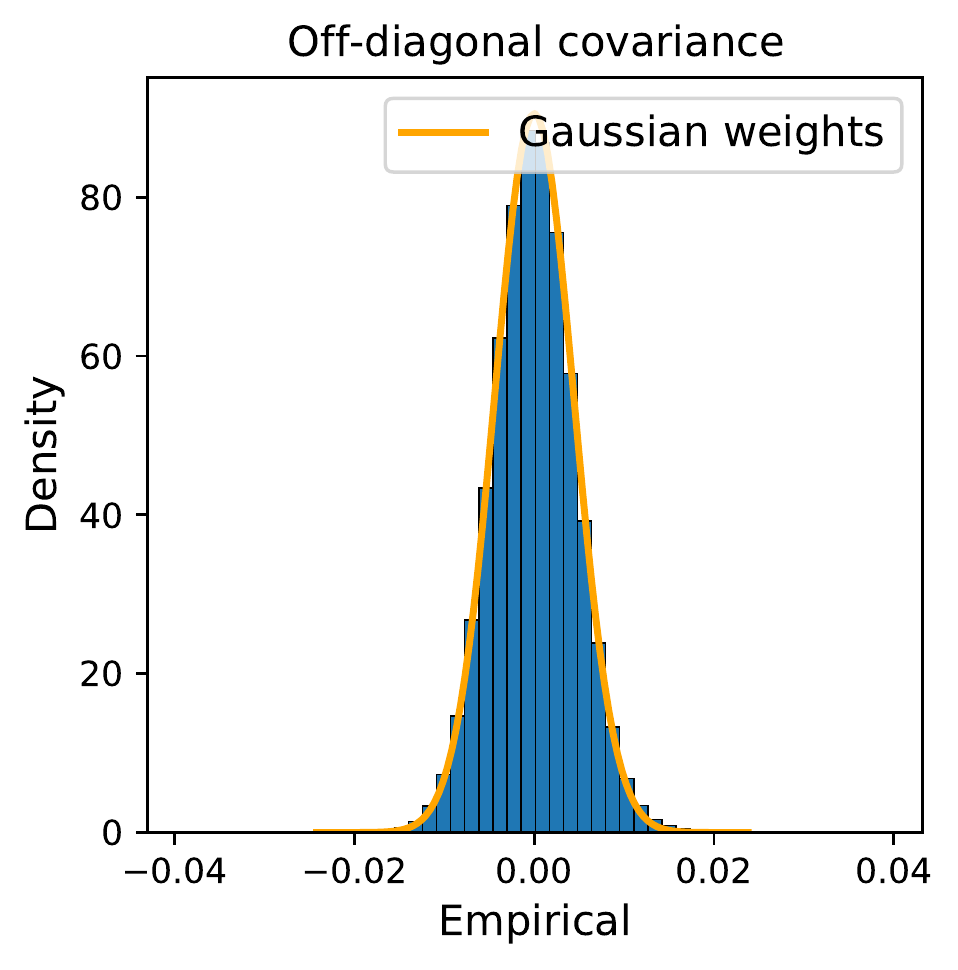}}
\end{tabular}
}
\caption{M1 off-diagonal covariance matrix value histograms}
\label{fig:m1_off_diag_covariance_elements}
\end{figure}

\begin{figure}[H]
\centering
\resizebox{\linewidth}{!}{
\begin{tabular}{ccc}
\subfloat[Attention queries]{\includegraphics[width=0.3\linewidth]{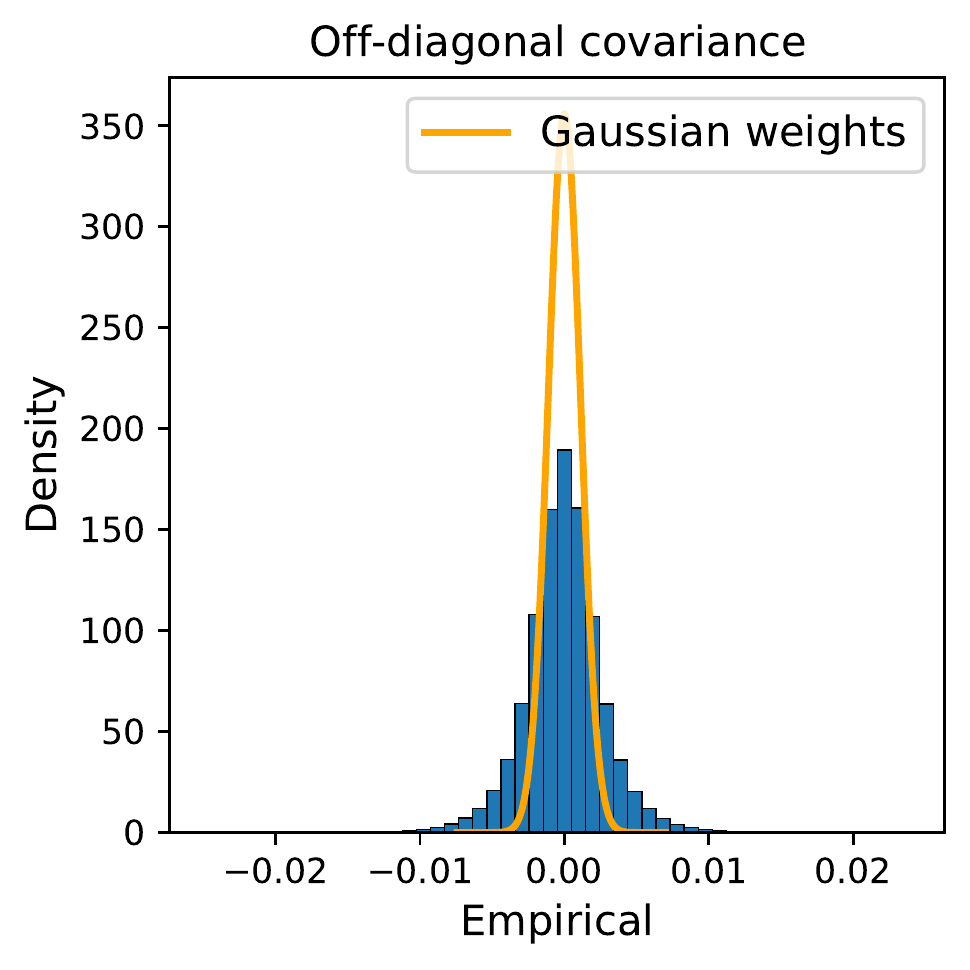}} &
\subfloat[Attention feedforward]{\includegraphics[width=0.3\linewidth]{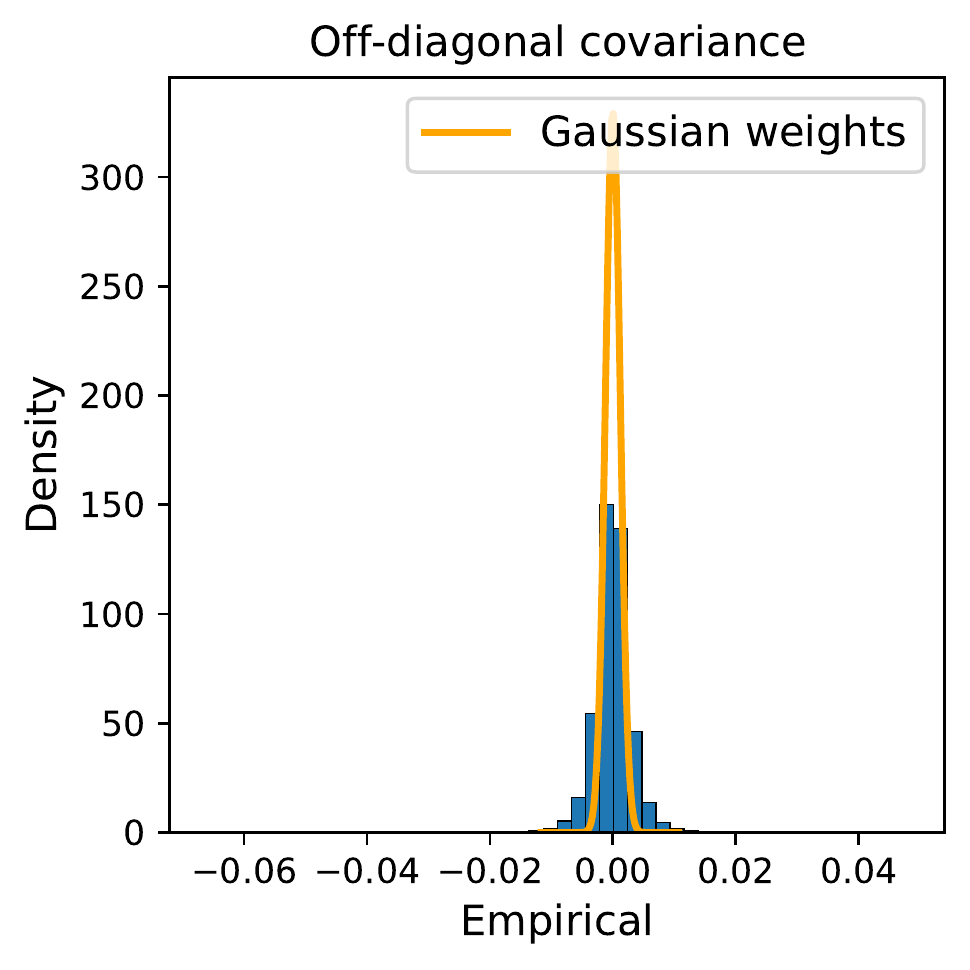}} &
\subfloat[Attention MLP hidden layer]{\includegraphics[width=0.3\linewidth]{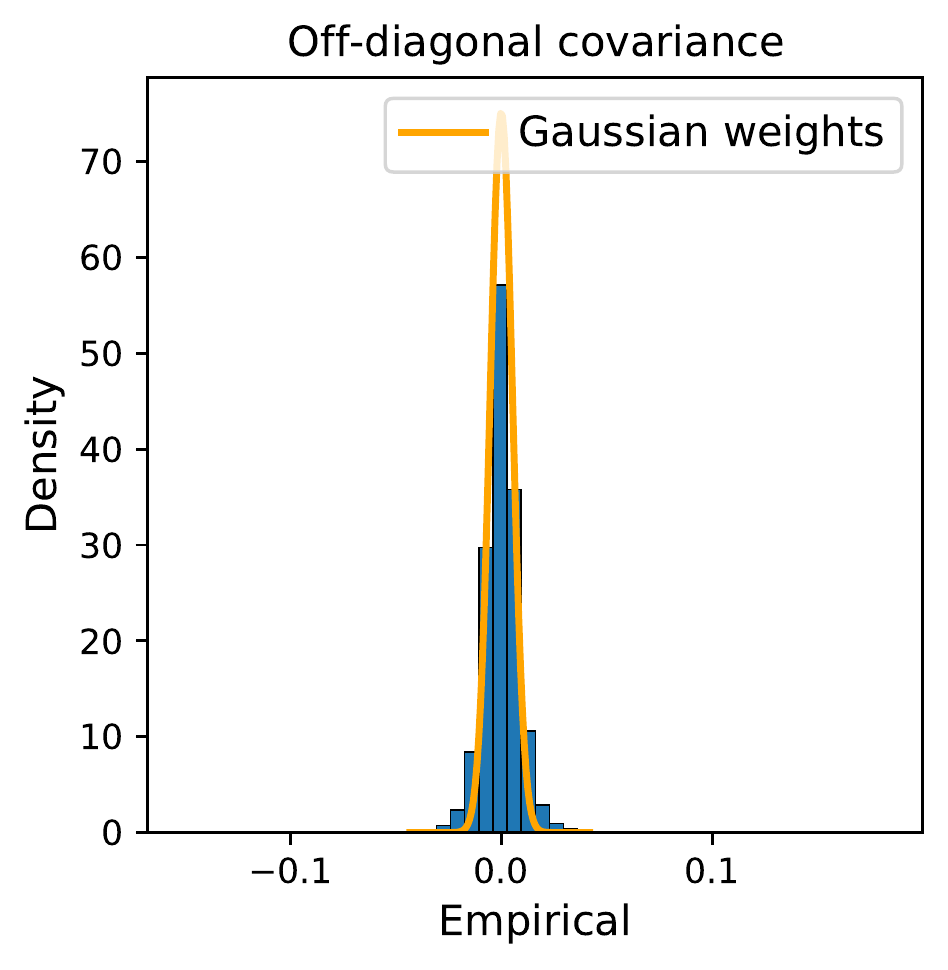}}
\end{tabular}
}
\caption{MNIST off-diagonal covariance matrix value histograms}
\label{fig:mnist_off_diag_covariance_elements}
\end{figure}

We here present histograms of the off-diagonal empirical covariance elements.  Covariance values have small magnitude and concentrate strongly in distribution around 0. Depending on the dataset, off-diagonal covariance elements are slightly larger than samples from an isotropic Gaussian as shown in Figure \ref{fig:m1_off_diag_covariance_elements} and \ref{fig:mnist_off_diag_covariance_elements}.

\begin{figure}[H]
\centering
\resizebox{\linewidth}{!}{
\begin{tabular}{ccc}
\subfloat[Attention queries]{\includegraphics[width=0.3\linewidth]{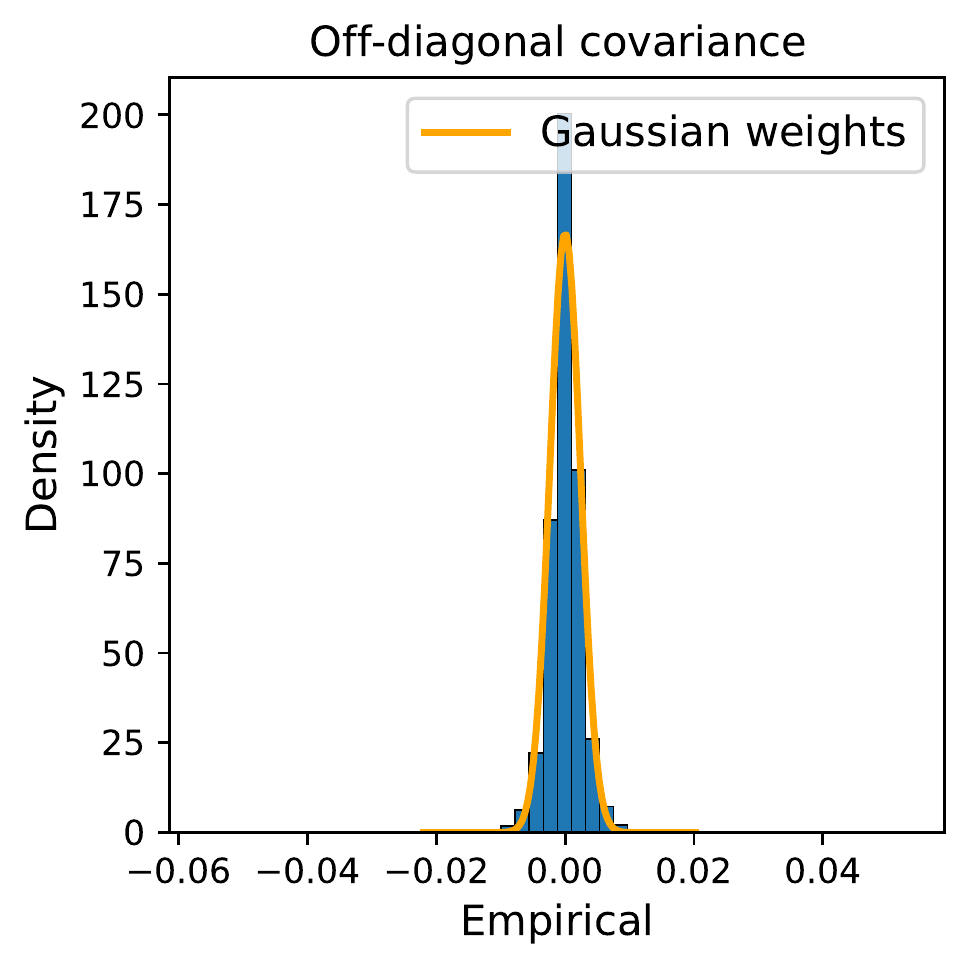}} &
\subfloat[Attention feedforward]{\includegraphics[width=0.3\linewidth]{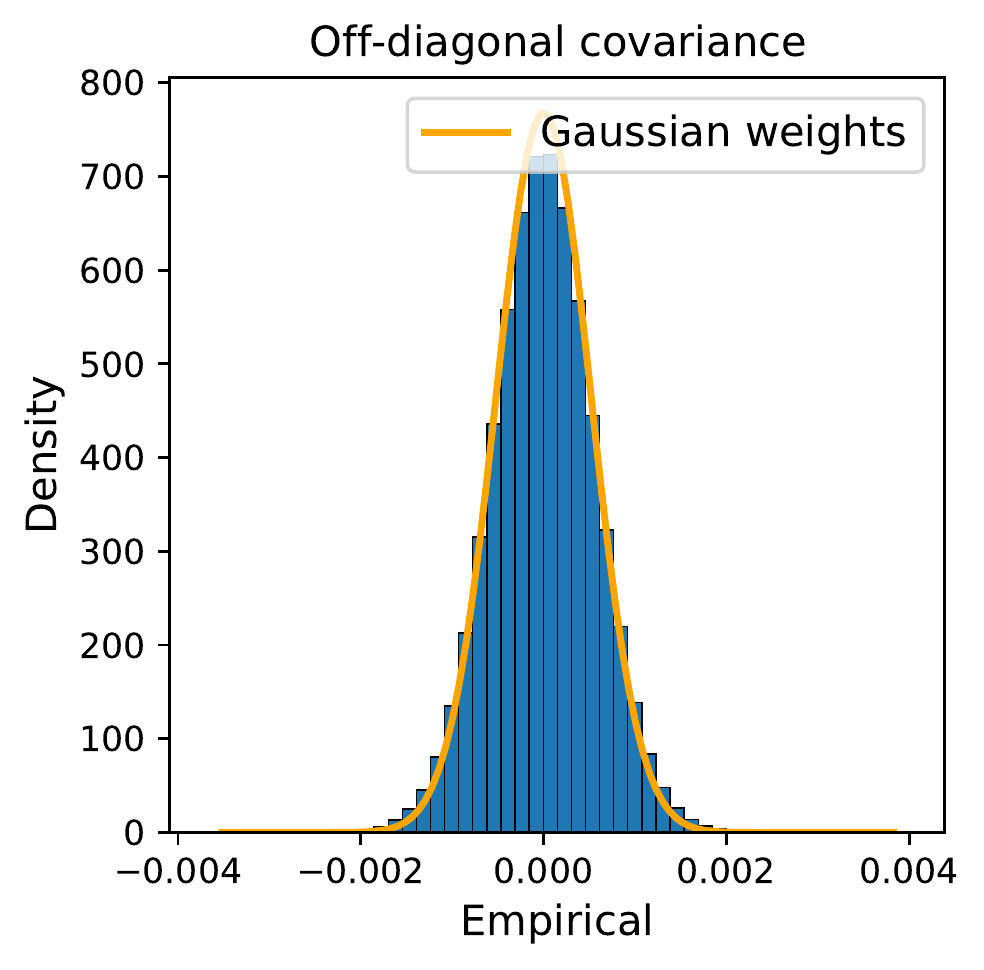}} &
\subfloat[Attention MLP hidden layer]{\includegraphics[width=0.3\linewidth]{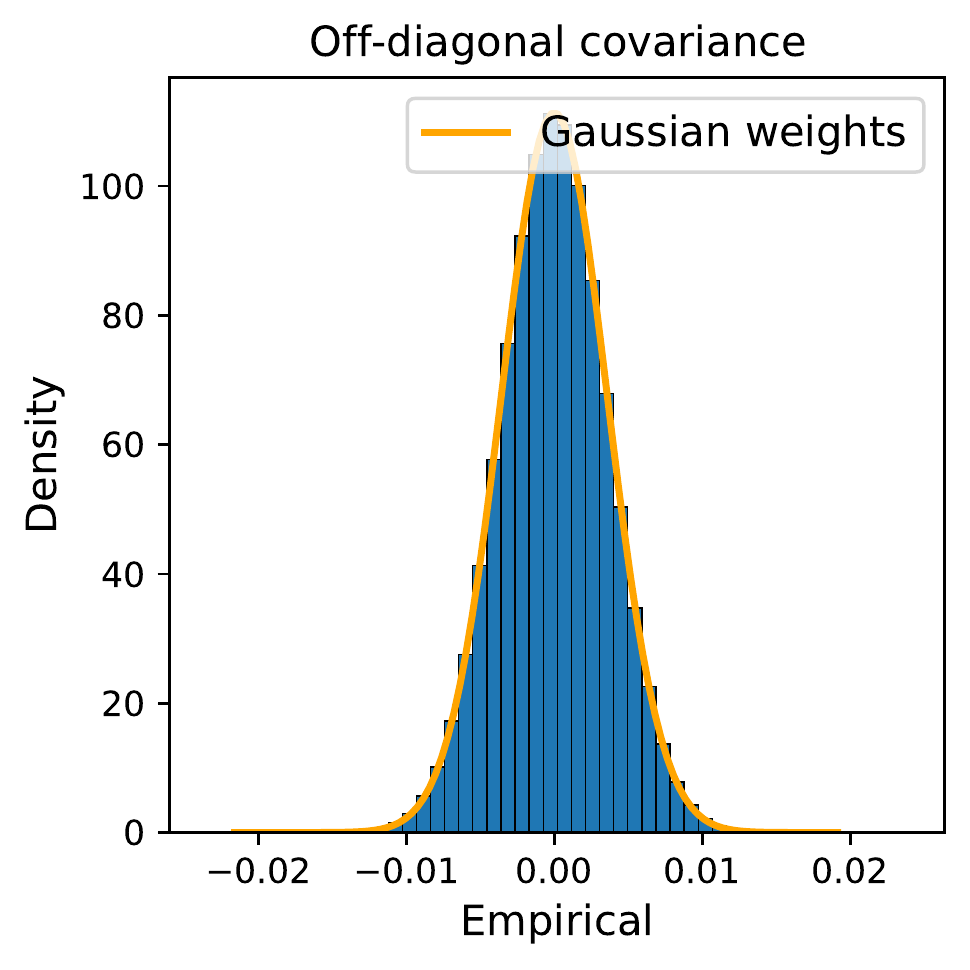}}
\end{tabular}
}
\label{fig:m2off_diag_covariance_elements}
\caption{M2 off-diagonal covariance matrix value histograms}
\end{figure}

\begin{figure}[H]
\centering
\resizebox{\linewidth}{!}{
\begin{tabular}{ccc}
\subfloat[Attention queries]{\includegraphics[width=0.3\linewidth]{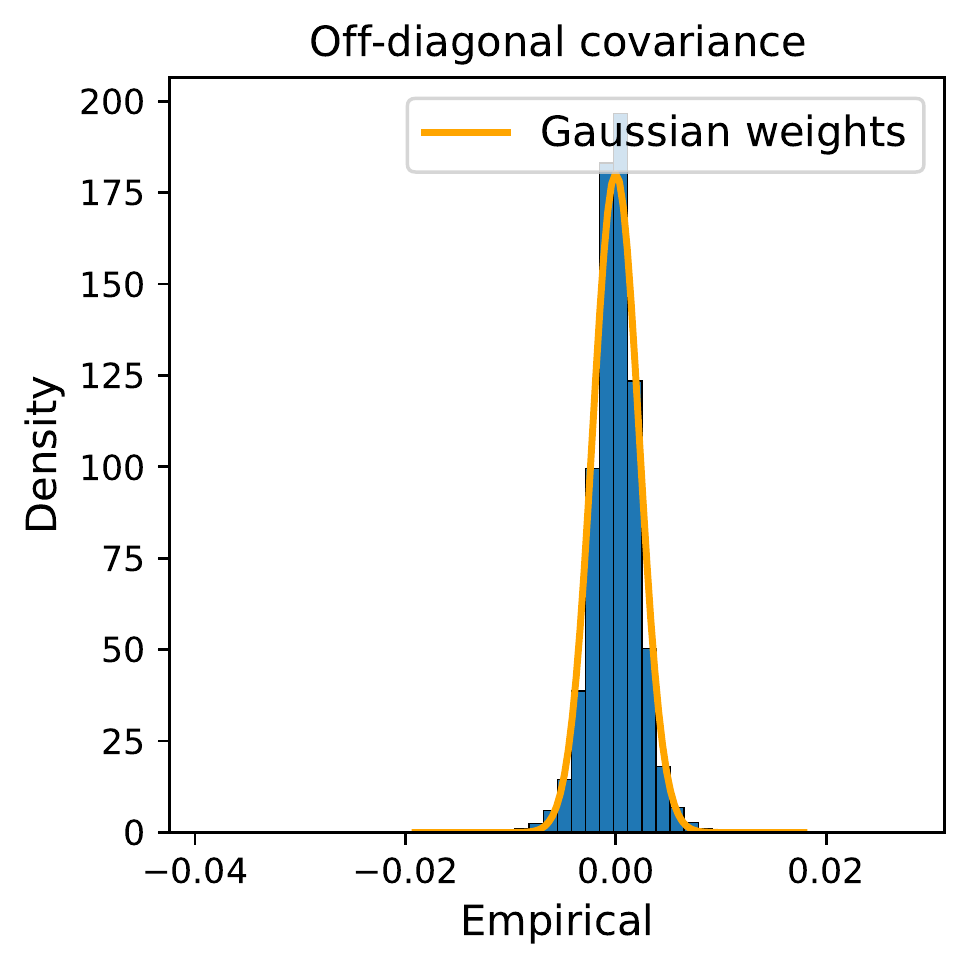}} &
\subfloat[Attention feedforward]{\includegraphics[width=0.3\linewidth]{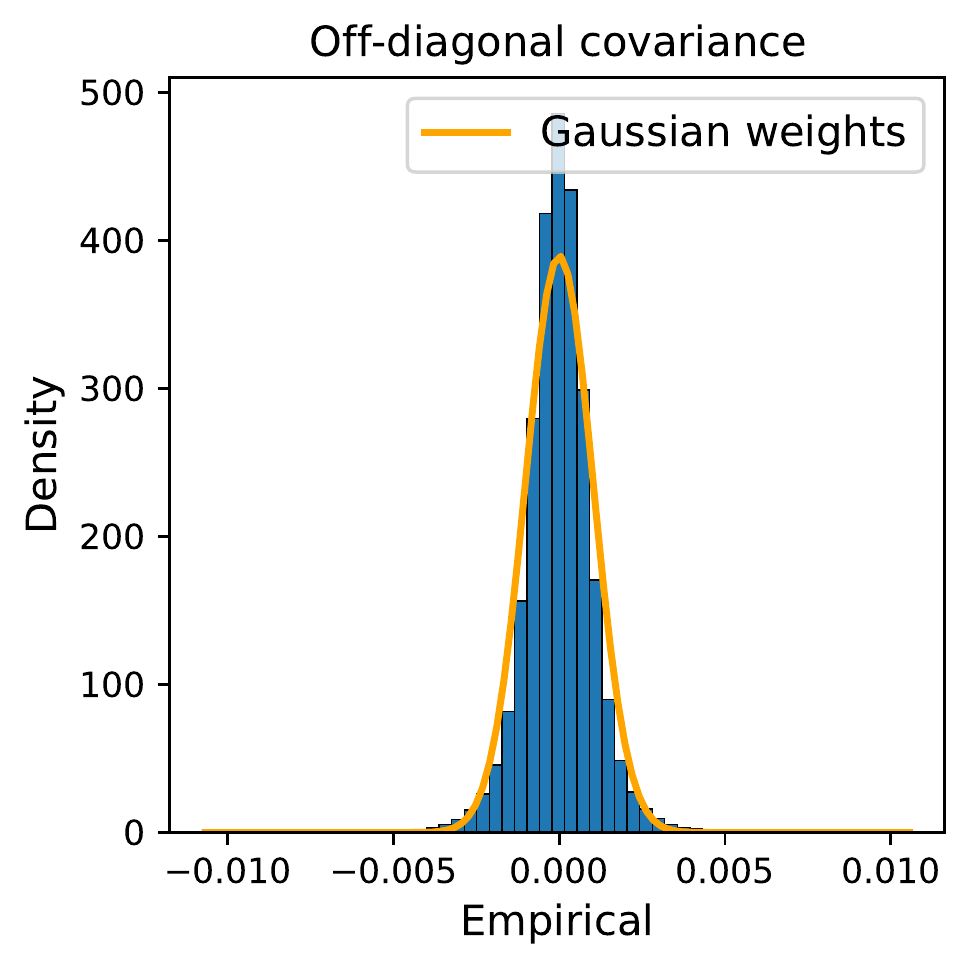}} &
\subfloat[Attention MLP hidden layer]{\includegraphics[width=0.3\linewidth]{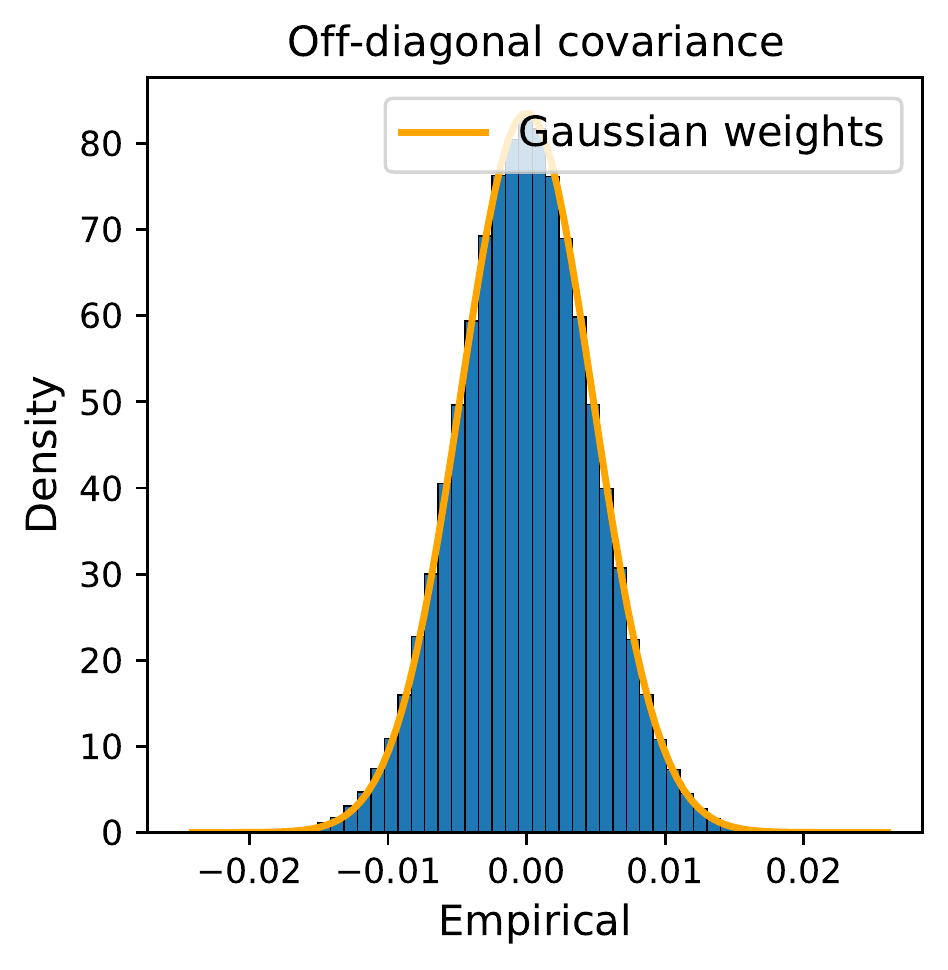}}
\end{tabular}
}
\caption{POS tagging off-diagonal covariance matrix value histograms}
\label{fig:pos_off_diag_covariance_elements}
\end{figure}

\subsubsection{Concrete dropout high/low entropy samples}

Here we present samples from the MNIST test set which have high (respectively low) predictive distribution entropy. By visual inspection, we find that high predictive entropy samples have an ambiguous labeling while low entropy predictive sample labels are much easier to identify.

\begin{figure}[H]
\resizebox{\linewidth}{!}{
\begin{tabular}{cc}
\centering
\subfloat[Highest entropy]{
    \captionsetup[subfigure]{labelformat=empty}
    \begin{tabular}{cccc}
        \subfloat[1\textsuperscript{st}]{\includegraphics[width=0.1\linewidth]{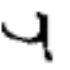}} &
        \subfloat[2\textsuperscript{nd}]{\includegraphics[width=0.1\linewidth]{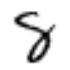}} &
        \subfloat[3\textsuperscript{rd}]{\includegraphics[width=0.1\linewidth]{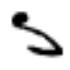}} & 
        \subfloat[4\textsuperscript{th}]{\includegraphics[width=0.1\linewidth]{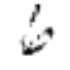}}
    \end{tabular}
}
& 
\subfloat[Lowest entropy]{
    \captionsetup[subfigure]{labelformat=empty}
    \begin{tabular}{cccc}
        \subfloat[1\textsuperscript{st}]{\includegraphics[width=0.1\linewidth]{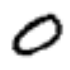}} &
        \subfloat[2\textsuperscript{nd}]{\includegraphics[width=0.1\linewidth]{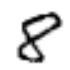}} &
        \subfloat[3\textsuperscript{rd}]{\includegraphics[width=0.1\linewidth]{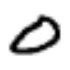}} & 
        \subfloat[4\textsuperscript{th}]{\includegraphics[width=0.1\linewidth]{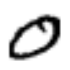}}
    \end{tabular}
}
\end{tabular}
}
\caption{Concrete Dropout : highest/lowest entropy MNIST test samples}
\end{figure}

\subsection{Implementation details}
\label{sec:imp_details}

In the following section, we present more detailed implementation specifications.

\subsubsection{Architecture \& Datasets}
    
\paragraph{Toy Data}  

We generate synthetic data from the two following models:
\begin{itemize}
    \item Model M1: $X_{t+1} = \sum_{i=0}^{4} 0.2 \cos(0.4\pi i X_t + 1/(i+1)) +\sqrt{0.5} \epsilon_{t+1}$
    \item Model M2: $X_{t+1} = \sum_{i=0}^{4} \sum_{j=0}^{4} 0.5 \cos(0.8\pi j X_{t-i}) + \sqrt{0.1} \epsilon_{t+1}$
\end{itemize}
where $(X_i)_{i=0}^{4}$ and $(\epsilon_t)_{t=1}^T$ are i.i.d standard Gaussian random variables.
We use 800 training, 80 validation and 80 testing sequences of length 24. For experiments with Toy Data, we use the transformer from \citep{vaswani2017attention} with GeLU activation \citep{hendrycks2020gelu}, a unique attention block, one attention head and a hidden size of 64. We train until convergence for 100 epochs. We draw 30 posterior samples to compute the predictive distribution and KL divergence.  We evaluate our model with the test data log-likelihood, mean squared error of the predicted variance and the expected mean square error given the previous value $X_t$.

\paragraph{Part of Speech Tagging}

We use the English split of the Part of Speech tagging dataset from the Universal Dependencies v1.2 corpus \citep{nivreuniversaldependencies}. This dataset contains 204'586 train, 25'148 validation and 25'096 test tokens. We use a maximum length of 40 tokens, pad the shorter sequences and split the sentences which exceed the text length. 
For experiments with this dataset, we use a transformer from \citep{vaswani2017attention} with GeLU activation, a unique attention block, one attention head and a hidden size of 32. We train until convergence for 100 epochs. We draw 10 posterior samples to compute the predictive distribution and KL divergence. We evaluate our model with the token level test data log-likelihood, token level accuracy, token level F1-score and expected calibration error (ECE) \citep{guo2017calibration}.

\paragraph{MNIST}

We experiment with the MNIST image classification dataset \citep{lecun2010mnist}. We split the original dataset into 48'000 training, 12'000 validation and 9'984 testing samples. 
For experiments with this dataset, we use a Vision transformer from \citep{dosovitskiy2021vit} with GeLU activation, two attention block, one attention head, a hidden size of 32 and a patch size of 4. We train until convergence for 150 epochs. We draw 10 posterior samples to compute the predictive distribution and KL divergence. We evaluate our model with the test data log-likelihood, accuracy, F1-score and ECE.

\subsubsection{Training setup}

Our models are trained using the Adam optimizer \citep{kingma2017adam} with the triangular learning rate schedule from \citep{vaswani2017attention}.
Our weight distribution experiments are conducted using 900 transformers trained by likelihood maximization with SGD as done in \citep{fortuin2021bayesian}. 
    
\subsubsection{Software packages}

We implement and train our models using the JAX \citep{jax2018github} and Haiku \citep{haiku2020github} Python libraries.

\end{document}